%% file: main.tex
\documentclass[twocolumn,final,10pt]{elsarticle}
\usepackage{medima}
\usepackage[english]{babel} 

\usepackage[full]{textcomp}
\renewcommand{\copyright}{\textcopyright}

\usepackage{mathdesign}
\usepackage{comment}
\usepackage{multirow}
\usepackage{booktabs}
\usepackage{subcaption}
\usepackage[justification=centering]{caption}
\usepackage[dvipsnames]{xcolor}
\usepackage{pgfplots}
\usepackage{soul}
\usetikzlibrary{calc}

\usepackage[normalem]{ulem}

\usepackage{makecell, cellspace, caption}

\usepackage{amssymb} 
\usepackage{amsfonts}
\usepackage{amsmath}
\usepackage{amsthm}
\usepackage{amsmath,amsfonts,amsthm,bm} 
\usepackage{hyperref} 
\hypersetup{
        colorlinks
}
\usepackage{algorithm} 
\usepackage{algpseudocode} 

\usepackage[switch,displaymath]{lineno}

\usepackage{graphicx}
\usepackage{times}
\usepackage{epsfig}
\usepackage{tabularx}
\usepackage{mathtools}
\usepackage{color, colortbl}
\usepackage{enumitem}
\usepackage{hhline}
\usepackage{bbm} 

\input{notation}

\journal{Medical Image Analysis}

\title{Calibrating Segmentation Networks with Margin-based Label Smoothing}

\algnewcommand\algorithmicinput{\textbf{Input:}}
\algnewcommand\Input{\item[\algorithmicinput]}
\algnewcommand\algorithmicoutput{\textbf{Output:}}
\algnewcommand\Output{\item[\algorithmicoutput]}

\verso{Murugesan \textit{et~al.}}

\author[1,2]{Balamurali \snm{Murugesan}\corref{cor1}}
\cortext[cor1]{Corresponding author: \url{balamurali.murugesan.1@ens.etsmtl.ca}}

\author[1,2]{Bingyuan \snm{Liu}}

\author[3]{Adrian \snm{Galdran}}

\author[1,2,4]{Ismail \snm{Ben Ayed}}

\author[1,2,4]{Jose \snm{Dolz}}

\address[1]{LIVIA, ÉTS Montréal, Canada}
\address[2]{International Laboratory on Learning Systems (ILLS),\\McGill - ETS - MILA - CNRS - Université Paris-Saclay - CentraleSupélec, Canada}
\address[3]{Universitat Pompeu Fabra, Barcelona, Spain}
\address[4]{Centre de Recherche du Centre Hospitalier de l'Université de Montréal (CRCHUM), Canada}

\received{1 September 2022}

\newcommand\und[1]{\underline{#1}}
\newcommand{\revf}[1]{{\color{black}#1}}
\newcommand{\rev}[1]{{\color{black}#1}}

\begin{document}
\begin{frontmatter}
\begin{abstract}
\rev{Despite the undeniable progress in visual recognition tasks fueled by deep neural networks, there exists recent evidence showing that these models are poorly calibrated, resulting in over-confident predictions. The standard practices of minimizing the cross-entropy loss during training promote the predicted softmax probabilities to match the one-hot label assignments. 
Nevertheless, this yields a pre-softmax activation of the correct class that is significantly larger than the remaining activations, which exacerbates the miscalibration problem.}
\rev{Recent observations from the classification} literature suggest that loss functions that embed implicit or explicit maximization of the entropy of predictions yield state-of-the-art calibration performances. \rev{Despite these findings, the impact of these losses in the relevant task of calibrating medical image segmentation networks remains unexplored.}
\rev{In this work,} we provide a unifying constrained-optimization perspective of current state-of-the-art calibration losses. Specifically, these losses could be viewed as approximations of a linear penalty (or a Lagrangian term) imposing equality constraints on logit distances. This points to an important limitation of such underlying equality constraints, whose ensuing gradients constantly push towards a non-informative solution, which might prevent from reaching the best compromise between the discriminative performance and calibration of the model during gradient-based optimization.
Following our observations, we propose a simple and flexible generalization based on inequality constraints, which imposes a controllable margin on logit distances. Comprehensive experiments on a variety of \rev{public medical image segmentation benchmarks} demonstrate that our method sets novel state-of-the-art results on these tasks in terms of network calibration, \rev{whereas the discriminative performance is also improved}. The code is available at \url{https://github.com/Bala93/MarginLoss} 
\end{abstract}

\end{frontmatter}

\setlength{\parskip}{3pt}

\section{Introduction}
\label{sec:intro}

\rev{Deep neural networks (DNNs) are driving progress in a variety of computer vision tasks across different domains and applications. In particular, these high-capacity models have become the \textit{de-facto} solution in critical tasks, such as medical image segmentation. Despite their superior performance, there exists recent evidence \citep{guo2017calibration,mukhoti2020calibrating,muller2019does} which demonstrates that these models are poorly calibrated, often resulting in over-confident predictions.} As a result, the predicted probability values associated with each class overestimate the actual likelihood of correctness.

\rev{Quantifying the predictive uncertainty of modern DNNs has gained popularity recently, with several alternatives to train better calibrated models. A simple yet effective approach consists in integrating a post-processing step that modifies the predicted probabilities of a trained neural network \citep{guo2017calibration,zhang2020mix,Tomani2021Posthoc,ding2021local}. This strategy, however, presents several limitations. First, the choice of the transformation parameters, such as temperature scaling, is highly dependent on the dataset and network. And second, under domain drift, post-hoc calibration performance largely degrades \citep{ovadia2019can}, resulting in unreliable predictions. A more principled alternative is to explicitly maximize the Shannon entropy of the model predictions during training, which can be achieved by augmenting the learning objective with a term that penalizes confident output distributions \citep{pereyra2017regularizing}.} Furthermore, recent efforts to quantify the quality of predictive uncertainties have focused on investigating the effect of the entropy on the training labels \citep{xie2016disturblabel,muller2019does,mukhoti2020calibrating}. Findings from these works evidence that, popular losses, which modify the hard-label assignments, such as label smoothing \citep{szegedy2016rethinking} and focal loss \citep{lin2017focal}, implicitly integrate an entropy maximization objective and have a favourable effect on model calibration. As shown comprehensively in the recent study in \citep{mukhoti2020calibrating}, these losses, with implicit or explicit maximization of the entropy, represent the state-of-the-art in model calibration in visual and non-visual recognition tasks.

Despite this progress, the benefit of these calibration losses remains unclear in medical image segmentation. Indeed, only a handful of works have addressed this important problem, mostly focusing on the calibration assessment of standard segmentation losses \citep{mehrtash2020confidence}, i.e., cross-entropy and Dice. \revf{From a clinical perspective, semantic segmentation is of pivotal importance in several downstream tasks, such as diagnostic, surgical planning, treatment assessment, or following-up disease progress. In these important steps, clinicians equipped with segmentation uncertainty can make better decisions, and build trust on the system. For example, a clinician faced with a large segmentation error localized in a particular area of an image and a small error at any other region, without knowledge of the segmentation uncertainty, may decide to dismiss the segmentation entirely. On the other hand, by providing precise estimates of the segmentation uncertainties, the clinician could evaluate whether these regions lie in low uncertainty or high uncertainty areas, which will facilitate the assessment of the quality of the segmentation per region. We stress that if clinicians place unwarranted confidence on regions with inaccurate uncertainty estimates, the resulting decision might have catastrophic consequences.} Thus, we believe that it is of great significance and interest to study methods for confidence calibration of segmentation models in the context of medical imaging.

\vspace{0.5cm}
\noindent The contributions of this work are summarized as follows: 

\begin{itemize}
    \item We provide a unifying constrained-optimization perspective of current state-of-the-art calibration losses. Specifically, these losses could be viewed as approximations of a linear penalty (or a Lagrangian term) imposing equality constraints on logit distances. This points to an important limitation of such underlying hard equality constraints, whose ensuing gradients constantly push towards a non-informative solution, which might prevent from reaching the best compromise between the discriminative performance and calibration of the model during gradient-based optimization.
    
    \item Following our observations, we propose a simple and flexible generalization based on inequality constraints, which imposes a controllable margin on logit distances.
    
    \item \rev{We provide comprehensive experiments and ablation studies on five different public segmentation benchmarks that focus on diverse targets and modalities, highlighting the generalization capabilities of the proposed approach. Our empirical results demonstrate the superiority of our method compared to state-of-the-art calibration losses in both calibration and discriminative performance.}
\end{itemize}

   \rev{This journal version provides a substantial extension of the conference work presented in \citep{liu2021devil}. In particular, we provide a thorough literature review on calibration of segmentation models, with a main focus on the medical field. Second, we perform a  comprehensive empirical validation, including \textit{i)} multiple public benchmarks covering diverse modalities and targets, \textit{ii)} adding recent approaches which specifically target calibration of segmentation models (i.e., \citep{islam2021spatially} and \citep{ding2021local}), and \textit{iii)} substantial in-depth analysis of the behaviour of the analyzed models. We believe that, to date, this work represents the most comprehensive evaluation of calibration models in the task of medical image segmentation, not only in terms of the amount of benchmarks employed, but also in regards of models compared. } 

\section{Related work}
\label{sec:related}

\noindent \textbf{Post-processing approaches.} \rev{Including a post-processing step that transforms the probability predictions of a deep network \citep{guo2017calibration, zhang2020mix, Tomani2021Posthoc, ding2021local} is a straightforward yet efficient strategy to mitigate miscalibrated predictions.} Among these methods, \textit{temperature scaling} \citep{guo2017calibration}, a variant of Platt scaling \citep{platt1999probabilistic}, employs a single scalar parameter over all the pre-softmax activations, which results in softened class predictions. Despite its good performance on in-domain samples, \citep{ovadia2019can} demonstrated that temperature scaling does not work well under data distributional shift. \citep{Tomani2021Posthoc} mitigated this limitation by transforming the validation set before performing the post-hoc calibration step, \rev{whereas \citep{Ma2021postrank} introduced a ranking model to improve the post-processing model calibration.}

\noindent \rev{\textbf{Probabilistic and non-probabilistic approaches} have been also investigated to measure the uncertainty of the predictions in modern deep neural networks.} For example, \rev{prior literature has employed Bayesian neural networks to approximate inference by learning a posterior distribution over the network parameters,} as obtaining the exact Bayesian inference is computationally intractable in deep networks. These Bayesian-based models include variational inference \citep{blundell2015weight,louizos2016structured}, stochastic expectation propagation \citep{hernandez2015probabilistic} or dropout variational inference \citep{gal2016dropout}. \rev{A popular non-parametric alternative is ensemble learning,} where the empirical variance of the network predictions is used as an approximate measure of uncertainty. This yields improved discriminative performance, as well as meaningful predictive uncertainty with reduced miscalibration. Common strategies to generate ensembles include differences in model hyperparameters \citep{wenzel2020hyperparameter}, random initialization of the network parameters and random shuffling of the data points \citep{lakshminarayanan2016simple}, Monte-Carlo Dropout \citep{gal2016dropout,zhang2019confidence}, dataset shift \citep{ovadia2019can} or model orthogonality constraints \citep{larrazabal2021orthogonal}. However, a main drawback of this strategy stems from its high computational cost, particularly for complex models and large datasets. 

\noindent \textbf{Explicit and implicit penalties.} Modern classification networks trained under the fully supervised learning paradigm resort to training labels provided as binary one-hot encoded vectors. Therefore, all the probability mass is assigned to a single class, resulting in minimum-entropy supervisory signals (i.e., entropy equal to zero). As the network is trained to follow this distribution, we are implicitly forcing it to be overconfident (i.e., to achieve a minimum entropy), thereby penalizing uncertainty in the predictions. While temperature scaling artificially increases the entropy of the predictions, \citep{pereyra2017regularizing} included into the learning objective a term to penalize confident output distributions by explicitly maximizing the entropy. In contrast to tackling overconfidence directly on the predicted probability distributions, recent works have investigated the effect of the entropy on the training labels. The authors of \citep{xie2016disturblabel} explored adding label noise as a regularization, where the disturbed label vector was generated by following a generalized Bernoulli distribution. Label smoothing \citep{szegedy2016rethinking}, which successfully improves the accuracy of deep learning models, has been shown to implicitly calibrate the learned models, as it prevents the network from assigning the full probability mass to a single class, while maintaining a reasonable distance between the logits of the ground-truth class and the other classes \citep{muller2019does}. 
More recently, \citep{mukhoti2020calibrating} demonstrated that focal loss \citep{lin2017focal} implicitly minimizes a Kullback-Leibler (KL) divergence between the uniform distribution and the softmax predictions, thereby increasing the entropy of the predictions. Indeed, as shown in \citep{muller2019does,mukhoti2020calibrating}, both label smoothing and focal loss implicitly regularize the network output probabilities, encouraging their distribution to be close to the uniform distribution. To our knowledge, and as demonstrated experimentally in the recent studies in \citep{muller2019does,mukhoti2020calibrating}, loss functions that embed implicit or explicit maximization of the entropy of the predictions yield state-of-the-art calibration performances.

\paragraph{\textbf{\rev{Calibration in medical image segmentation}}}\rev{Recent literature has focused on either estimating the predictive uncertainty or on leveraging this uncertainty to improve the discriminative performance of segmentation models \citep{wang2019aleatoric}. Nevertheless, research to improve both the calibration and segmentation performance of CNN-based segmentation models is scarce. 
\citep{jena2019bayesian} proposed a novel deep segmentation framework rooted in generative modeling and Bayesian decision theory, which allowed to define a principled measure of uncertainty associated with label probabilities. Recent findings \citep{fort2019deep}, however, suggest that current state-of-the-art Bayesian neural networks have tendency to find solutions around a single minimum of the loss landscape and, consequently, lack diversity. 
In contrast, ensembling deep neural networks typically results in more diverse predictions, and therefore obtain better uncertainty estimates. This observation aligns with the recent work in \citep{jungo2020analyzing,mehrtash2020confidence}, which evaluates several uncertainty estimation approaches and concludes that ensembling outperforms other methods. 
To promote model diversity within the ensemble, \citep{larrazabal2021orthogonal} integrate an orthogonality constraint in the learning objective, showing significant gains over the non-constrained set. More recently, \citep{karimi2022improving} argue that training a single model in a multi-task manner on several different datasets yields better calibration on the different tasks \revf{compared to its single-task counterpart}. Nevertheless, these methods incur in high computationally expensive steps as they involve training either multiple models or a single model on multiple datasets. In an orthogonal direction, several recent methods have overcome this limitation and proposed lighter alternatives. For example, \citep{ding2021local} extends the naive temperature scaling by integrating a simple CNN to predict the pixel-wise temperature values in a post-processing step. \revf{Despite the improvement observed over the naive TS, this method inherits the limitations of temperature scaling and related post-processing approaches.} 
In addition, \citep{islam2021spatially} apply a weight matrix with a Gaussian kernel across the one-hot encoded expert labels to obtain soft class probabilities, adding into the standard Label smoothing a spatial-awareness. \revf{A potential limitation of this strategy is that the modification of the hard labels is done without considering the behaviour of the model, systematically disregarding those samples which are more, or less, confident. This contrasts with the proposed approach, which does not modify the hard assignments, but directly controls the confidence of the model in the logit space.} 
However, despite these initial efforts, and to the best of our knowledge, a comprehensive evaluation of calibration methods in multiple medical image segmentation benchmarks has not been conducted yet.}

\section{Preliminaries}
\label{sec:form}

\rev{Let $\mathcal{D}(\mathcal{X}, \mathcal{Y})=\{(\xx^{(i)}, \yy^{(i)})\}_{i=1}^N$ be the training dataset, with $\xx^{(i)} \in \mathcal{X} \subset \mathbb{R}^{\Omega_i}$ representing the $i^{th}$ image,} $\Omega_i$ the spatial image domain, and $\yy \in \mathcal{Y} \subset \mathbb{R}^K$ its corresponding ground-truth label with $K$ classes, provided as one-hot encoding. 
Given an input image $\xx^{(i)}$, a neural network parameterized by $\theta$ generates a logit vector, defined as $f_{\theta}(\xx^{(i)})=\mathbf{l}^{(i)} \in \mathbb{R}^K $. To simplify the notations, we omit sample indices, as this does not lead to ambiguity, and just use $\mathbf{l} = (l_k)_{1 \leq k \leq K} \in \mathbb{R}^K$ to denote logit vectors. Note that the logits are the inputs of the softmax probability predictions of the network, which are computed
as:
\[\sss = (s_k)_{1 \leq k \leq K} \in \mathbb{R}^K; \quad s_{k} = \frac{\exp^{l_k}}{\sum_j^K \exp^{l_j}}\]
The predicted class is computed as $\hat{y} = \argmax_k s_k$, whereas the predicted confidence is given by $\hat{p} = \max_{k} s_k$.

\noindent \textbf{Calibrated models.} 
\textit{Perfectly calibrated} models are those for which the predicted confidence for each sample is equal to the model accuracy : $\hat{p} = \mathbb{P}(\hat{y} = y | \hat{p})$, where $y$ denotes the true labels.
Therefore, an \textit{over-confident model} tends to yield predicted confidences that are larger than its accuracy, whereas an \textit{under-confident model} displays lower confidence than the model's accuracy.


\noindent \textbf{Miscalibration of DNNs.} To train fully supervised discriminative deep models, the standard cross-entropy (CE) loss is commonly used as the training objective. We argue that, from a calibration performance, the supervision of CE is suboptimal. Indeed, CE reaches its minimum when the predictions for all the training samples match the hard (binary) ground-truth labels, i.e., $s_k = 1$ when $k$ is the ground-truth class of the sample and $s_k = 0$ otherwise. Minimizing the CE implicitly pushes softmax vectors $\sss$ towards the vertices of the simplex, thereby magnifying the distances between the largest logit $\max_k(l_k)$ and the rest of the logits, yielding over-confident predictions and miscalibrated models. 

\section{A constrained-optimization perspective of calibration}
\label{sec:view}

We present in this section a novel constrained-optimization perspective of current calibration methods for deep networks, showing that the existing 
strategies, including Label Smoothing (LS) \citep{muller2019does,szegedy2016rethinking}, Focal Loss (FL) \citep{mukhoti2020calibrating,lin2017focal} and Explicit Confidence Penalty (ECP) \citep{pereyra2017regularizing}, impose {\em equality} constraints on logit distances. Specifically, they embed either explicit or implicit penalty functions, which push all the logit distances to zero.

\subsection{Definition of logit distances}
Let us first define the vector of logit distances between the \revf{winner class (i.e., the class with the highest logit: $\argmax_j (l_j)$)} and the \revf{remaining classes} as:
\begin{align}\label{eq:logit distance}
    \dd (\llll) = (\max_j (l_j) - l_k)_{1 \leq k \leq K} \in \mathbb{R}^{K} 
\end{align}
Note that each element in $\dd(\llll)$ is non-negative.
In the following, we show that LS, FL and ECP correspond to different {\em soft penalty} functions for imposing the same hard equality constraint $\dd (\llll) = \mathbf 0$ or, equivalently, imposing inequality 
constraint $\dd (\llll) \leq \mathbf 0$ (as $\dd (\llll)$ is non-negative by definition). 
Clearly, enforcing this equality constraint in a hard manner would result in all $K$ logits being equal for a given sample, which corresponds to non-informative softmax predictions $s_k = \frac{1}{K} \, \forall k$.  

\subsection{Penalty functions in constrained optimization}

In the general context of constrained optimization \citep{Bertsekas95}, {\em soft} penalty functions are widely used to tackle {\em hard} equality or inequality constraints. For the discussion in the sequel, consider specifically the following hard equality constraint:
\begin{equation}
\label{equality-constraint-label-smoothing}
\dd (\llll) = {\mathbf 0} 
\end{equation}
The general principle of a soft-penalty optimizer is to replace a hard constraint of the form in Eq.~\ref{equality-constraint-label-smoothing} by adding an additional term $\mathcal{P}(\dd (\llll))$ into the main objective function to be minimized. Soft penalty $\mathcal{P}$ should be a continuous and differentiable function, which reaches its global minimum when the constraint is satisfied, i.e., it verifies: $\mathcal{P}(\dd (\llll)) \geq \mathcal{P}(\mathbf {0}) \, \forall \, \llll \in \mathbb{R}^{K}$.
Thus, when the constraint is violated, i.e., when $\dd (\llll)$ deviates from $\mathbf {0}$, the penalty term $\mathcal{P}$ increases.

\noindent \textbf{Label smoothing.} \rev{Recent evidence \citep{lukasik2020does,muller2019does} suggests that, in addition to improving the discriminative performance of deep neural networks, Label Smoothing (LS) \citep{szegedy2016rethinking} positively impacts model calibration. }In particular, LS modifies the hard target labels with a smoothing parameter $\alpha$, so that the original one-hot training labels $\yy \in \{0, 1\}^K$ become $\yy^\text{LS} = (y_{k}^\text{LS})_{1 \leq k \leq K}$, with $y_{k}^\text{LS}=y_{k}(1-\alpha)+\frac{\alpha}{K}$. Then, we simply minimize the cross-entropy between the modified labels and the network outputs:
\begin{align}
\label{eq:ls}
    {\cal L}_\text{LS} = -\sum_{k} y_{k}^\text{LS} \log s_{k} = -\sum_{k} ((1-\alpha)y_k + \frac{\alpha}{K}) \log s_{k}
\end{align}
where $\alpha \in [0,1]$ is the smoothing hyper-parameter.
It is straightforward to verify that cross-entropy with label smoothing in Eq.~\ref{eq:ls} can be decomposed into a 
standard cross-entropy term augmented with a Kullback-Leibler (KL) divergence between uniform distribution ${\mathbf u} = \frac{1}{K}$ and the softmax prediction:
\begin{align}
\label{eq:ls-kl}
{\cal L}_\text{LS} \ceq {\cal L}_\text{CE} + \frac{\alpha}{1-\alpha}{\cal D}_\text{KL}\left({\mathbf u} || \s \right )
\end{align}
where $\ceq$ stands for equality up to additive and/or non-negative multiplicative constants. Now, consider the following bounding relationships between a linear penalty (or a Lagrangian) for equality constraint $\dd (\llll) = {\mathbf 0}$ and the KL divergence in Eq.~\ref{eq:ls-kl}.  
\begin{proposition}
\label{prop:ls}
A linear penalty (or a Lagrangian term) for constraint $\dd (\llll) = {\mathbf 0}$ is bounded from above and below
by ${\cal D}_\text{KL}\left({\mathbf u} || \s \right )$, up to additive constants:
\begin{align}
{\cal D}_\text{KL}\left({\mathbf u} || \s \right ) - \log(K) \cleq \frac{1}{K}\sum_k (\max_j (l_j) - l_k) \cleq  {\cal D}_\text{KL}\left({\mathbf u} || \s \right ) \nonumber
\end{align}
where $\cleq$ stands for inequality up to an additive constant.
\end{proposition}
These bounding relationships could be obtained directly from the softmax and ${\cal D}_\text{KL}$
expressions, along with the following well-known property of the LogSumExp function: $\max_k(l_k) \leq \log{\sum_k^K e^{l_k}} \leq \max_k(l_k) + \log(K)$.
For the details of the proof, please refer to Appendix A \rev{of the conference version in \citep{liu2021devil}}.

\begin{figure}[h]
    \centering
    \includegraphics[width=1\columnwidth]{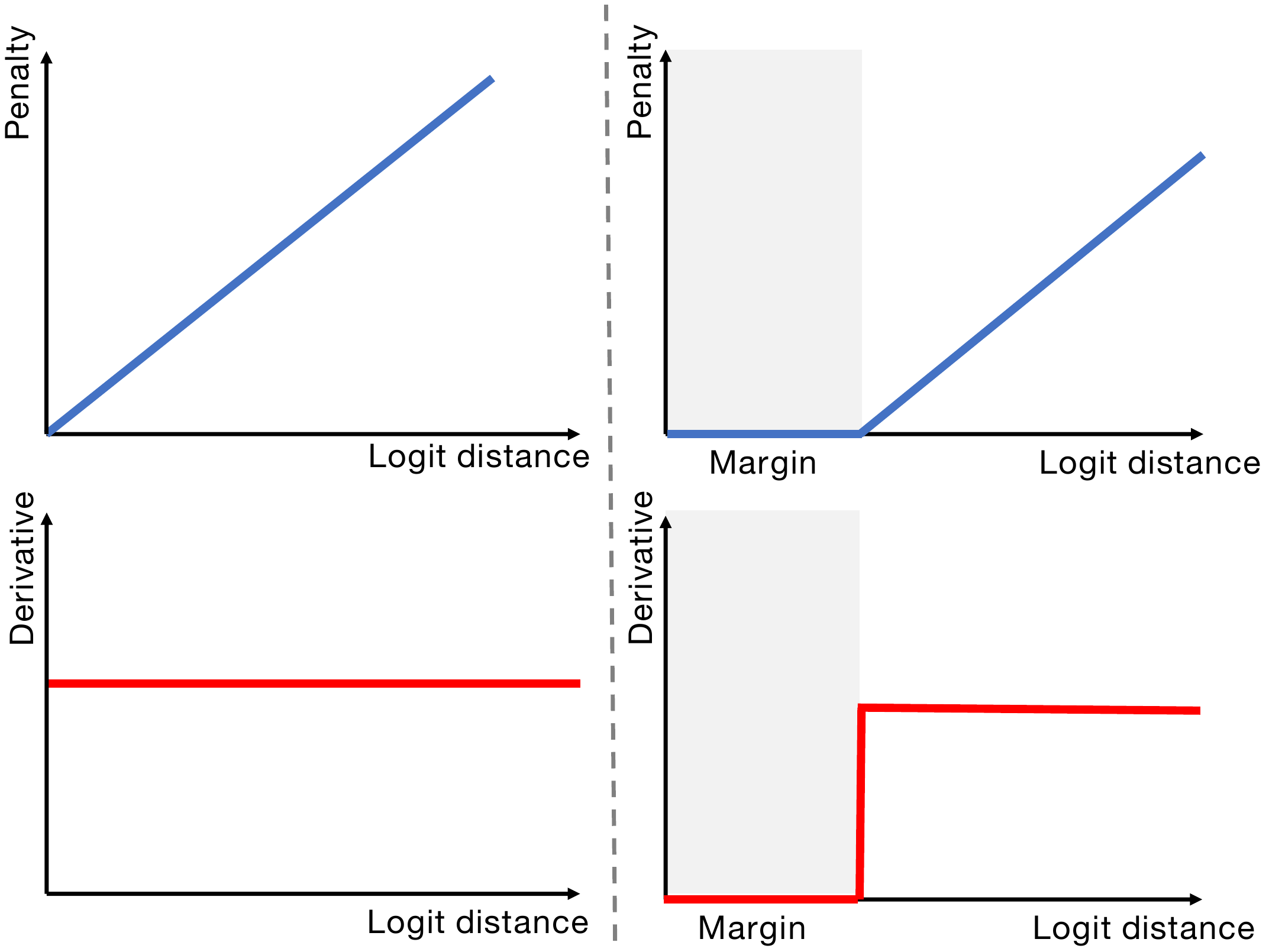}
    \caption{Illustration of the linear (left) and margin-based (right) penalties for imposing 
    logit-distance constraints, along with the corresponding derivatives. \rev{Note that while the derivative of the linear penalty for constraint $\mathbf{d}(\mathbf{l})=\mathbf{0}$ constantly pushes towards the trivial solution $s_k=\frac{1}{K}\forall K$ (i.e., LS, FL and EPC), the derivative of the proposed model only pushes towards zero those logits above the given margin.}}
    \label{fig:method}
    \vspace{-2mm}
\end{figure}

Prop. \ref{prop:ls} means that LS is (approximately) optimizing a linear penalty (or a Lagrangian) for logit-distance constraint $\dd(\llll) = \0$, which encourages equality of all logits; see the illustration in Figure~\ref{fig:method}, top-left.  

\noindent \textbf{Focal loss.} Another popular alternative for calibration is focal loss (FL) \citep{lin2017focal}, which attempts to alleviate the over-fitting issue in CE by directing the training attention towards samples with low confidence in each mini-batch. More concretely, the authors 
proposed to use a modulating factor to the CE, $(1-s_k)^{\gamma}$, which controls the trade-off between easy and hard examples. Very recently, \citep{mukhoti2020calibrating} demonstrated that focal loss is, in fact, an upper bound on CE augmented with a term that implicitly serves as a maximum-entropy regularizer:
\begin{align}
{\cal L}_\text{FL} = -\sum_{k} (1 - s_{k})^{\gamma} y_{k} \log s_{k} \geq \mathcal{L}_{\text{CE}} - \gamma\mathcal{H}(\mathbf{s})
\label{eq:fl}
\end{align}
where $\gamma$ is a hyper-parameter and $\mathcal{H}$ denotes the Shannon entropy of the softmax prediction, given by
\[\mathcal{H}(\mathbf{s}) = -\sum_k s_k \log(s_k)\]
In this connection, FL is closely related to ECP \citep{pereyra2017regularizing}, which explicitly added the negative entropy term, $-\mathcal{H}(\mathbf{s})$, to the training objective. It is worth noting that minimizing the negative entropy of the prediction is equivalent to minimizing the KL divergence between the prediction and the uniform distribution, up to an additive constant, i.e., 
\[-\mathcal{H}(\mathbf{s}) \ceq {\cal D}_\text{KL}(\s || \mathbf{u})\] 
which is a reversed form of the KL term in Eq.~\ref{eq:ls-kl}. 

Therefore, all in all, and following Prop.~\ref{prop:ls} and the discussions above, 
LS, FL and ECP could be viewed as different penalty functions for imposing the same logit-distance equality constraint $\dd(\llll) = \0$. This motivates our margin-based generalization of logit-distance constraints, which we introduce in the following section, along with discussions of its desirable properties (e.g., gradient dynamics) for calibrating neural networks. 

\subsection{Margin-based Label Smoothing (MbLS)}
\label{sec:our}

Our previous analysis shows that LS, FL and ECP are closely related from a constrained-optimization perspective, and they could be seen as approximations of a linear penalty for imposing constraint $\dd (\llll) = {\mathbf 0}$, pushing all logit distances to zero; see Figure~\ref{fig:method}, top-left. Clearly, enforcing this constraint in a hard way yields a non-informative solution where all the classes have exactly the same logit and, hence, the same class prediction: $s_k = \frac{1}{K}\, \forall K$. While this trivial solution is not reached in practice when using soft penalties (as in LS, FL and ECP) jointly with CE, we argue that the underlying equality constraint $\dd (\llll) = {\mathbf 0}$ has an important limitation, which might prevent from reaching the best compromise between the discriminative performance and calibration of the model during gradient-based optimization. Figure \ref{fig:method}, left, illustrates this: With the linear penalty for constraint $\dd (\llll) = {\mathbf 0}$ in the top-left of the Figure, the 
derivative with respect to logit distances is a strictly positive constant (left-bottom), yielding during training {\em a gradient term that constantly pushes towards the trivial, non-informative solution} $\dd (\llll) = {\mathbf 0}$ (or equivalently $s_k = \frac{1}{K}\, \forall K$). To alleviate this issue, we propose to replace the
equality constraint $\dd (\llll) = {\mathbf 0}$ with the more general inequality constraint $\dd (\llll) \leq {\mathbf m} $, where ${\mathbf m}$ denotes the $K$-dimensional vector with all elements equal to $m > 0$.
Therefore, we include a margin $m$ into the penalty, so that the logit distances in $\dd(\llll)$ are allowed to 
be below $m$ when optimizing the main learning objective:
\begin{align}
\label{eq:our-constraint}
    \min \quad   \mathcal{L}_{\text{CE}} \quad \text{s.t.} \quad  \dd(\llll) \leq \textbf{m} , \quad \textbf{m} > \textbf{0}
\end{align}
The intuition behind adding a strictly positive margin $m$ is that, unlike the linear penalty for constraint $\dd (\llll) = {\mathbf 0}$ (Figure \ref{fig:method}, left), the gradient is back-propagated only on those logits where the distance is above the margin (Figure \ref{fig:method}, right). This contrasts with the 
linear penalty, for which there exists always a gradient, and its value is the same across all the logits, regardless of their distance.

Even though the constrained problem in Eq. \ref{eq:our-constraint} could be solved by a Lagrangian-multiplier algorithm, we resort to a simpler unconstrained approximation by ReLU function:
\begin{align}
\label{eq:our-l1}
    \min \quad &  \mathcal{L}_{\text{CE}} + \lambda \sum_k \max(0, \max_j (l_j) - l_k - m)
\end{align}
Here, the non-linear ReLU penalty for inequality constraint $\dd (\llll) \leq {\mathbf m}$ discourages logit distances from surpassing a given margin $m$, and $\lambda$ is a trade-off weight balancing the two terms.
It is clear that, as discussed in Sec.~\ref{sec:view}, several competitive calibration methods could be viewed as approximations for imposing constraint $\dd (\llll) = {\mathbf 0}$ and, therefore, correspond to the special case of our method when setting the margin to $m=0$. Our comprehensive experiments in the next section demonstrate clearly the benefits of introducing a strictly positive margin $m$.

Note that our model in Eq.~\ref{eq:our-l1} has two hyper-parameters, $m$ and $\lambda$. We fixed $\lambda$ to $0.1$ in our experiments for all the benchmarks, and tuned only the margin $m$ over validation sets. In this way, when comparing with the existing calibration solutions, we use the same budget of hyper-parameter optimization ($m$ in our method vs. $\alpha$ in LS or $\gamma$ in FL).

\section{Experiments}

\subsection{\rev{Experimental Setting}}

\subsubsection{\rev{Datasets}}
\rev{To empirically validate our model, we employ five public multi-class segmentations benchmarks, whose detailes are specified below. 
} 

\paragraph{\textbf{\rev{Automated Cardiac Diagnosis Challenge (ACDC)}} \citep{bernard2018deep}} \rev{This dataset consists of 100 patient exams containing cardiac MR volumes and its respective multi-class segmentation masks for both diastolic and systolic phases. The segmentation mask contains four classes, including the left ventricle (LV), right ventricle (RV), myocardium (Myo) and background. Following the standard practices on this dataset, 2D slices are extracted from the available volumes and resized to 224$\times$224. Last, the dataset is randomly split into independent training (70), validation (10) and testing (20) sets.}

\paragraph{\textbf{\revf{Brain Tumor Segmentation (BRATS) 2019 Challenge}}} \citep{Menze2015TheBRATSJ, Bakas2017AdvancingFeaturesJ, Bakas2018IdentifyingChallengeJ} \revf{The dataset contains $335$ multi-modal MR scans (FLAIR, T1, T1-contrast, and T2) with their corresponding Glioma segmentation masks. The classes representing the mask include tumor core (TC), enhancing tumor (ET) and whole tumor (WT). Each volume of dimension 155$\times$240$\times$240 is resampled and slices containing only background are removed from the training. The patient volumes are randomly split to 235, 35, 65 for training, validation, and testing, respectively.}

\paragraph{\textbf{\rev{MRBrainS18}} \citep{mrbrains_dataset}} \rev{The dataset contains paired T1, T2, and T1-IR volumes of 7 subjects and their segmentation masks, which correspond to brain tissue including Gray Matter (GM), White Matter (WM), and Cerebralspinal fluid (CSF). The dimensions of the volumes are 240$\times$240$\times$48. We utilize 5 subjects for training and 2 subjects for testing.} 

\paragraph{\textbf{\rev{Fast and Low GPU memory Abdominal oRgan sEgmentation (FLARE) Challenge \citep{AbdomenCT-1K}}}} \rev{The dataset contains $360$ volumes of multi-organ abdomen CT including liver, kidneys, spleen and pancreas and their corresponding pixel-wise masks. The different resolutions are resampled to a common space and cropped to 192$\times$192$\times$30. The volumes are then randomly split to 240 for training, 40 for validation, 80 for testing.}

\paragraph{\textbf{\rev{PROMISE}}} \citep{litjens2014evaluation} \rev{The dataset was made available at the MICCAI 2012 prostate MR segmentation challenge. It contains the transversal T2-weighted MR images acquired at different centers with multiple MRI vendors and different scanning protocols. It is comprised of various diseases, i.e., benign and prostate cancers. The images resolution ranges from 15$\times$256$\times$256 to 54$\times$512$\times$512 voxels with a spacing ranging from 2$\times$0.27$\times$0.27 to 4$\times$0.75$\times$0.75mm$^3$. We employed $22$ patients for training, $3$ for validation and $7$ for testing.}

\paragraph{\textbf{\revf{HIPPOCAMPUS (HPC)}}} \citep{antonelli2022medical} \revf{: The data set consists of 260 MRI images acquired at the Vanderbilt University Medical Center, Nashville, US. This data set was selected due to the precision needed to segment such a small object in the presence of a complex surrounding environment. T1-weighted MPRAGE was used as the imaging sequence. The corresponding target ROIs were the anterior and posterior of the hippocampus, defined as the hippocampus proper and parts of the subiculum.  The data is split to 185, 25, 50 for training, validation, and testing, respectively.} 

\paragraph{\textbf{\revf{Breast UltraSound Images (BUSI)}}} \citep{al2020dataset} \revf{The datasets consists of ultrasound images of normal, benign and malignant cases of breast cancer along with the corresponding segmentation maps. We use only benign and mailgnant images, which results in a total of 647 images resized to a resolution of 256 $\times$ 256 to benchmark our results. We considered 445 images for training, 65 images for validation, and the remaining 137 images for testing. }

Note that in all datasets, images are normalized to be within the range [0-1]. Furthermore, for the datasets containing multiple image modalities (i.e., MRBrainS and BRATS), all available modalities are concatenated in a single tensor, which is fed to the input of the neural network. In addition, there exists one dataset for which the low amount of available images impeded us to generate a proper training, validation and testing split (MRBrainS). \revf{In this case, we performed leave-one-out-cross-validation in our experiments, whereas the other datasets followed standard training, validation and testing procedures, using a single split in the experiments.}

\rev{To assess the discriminative performance of the evaluated models, we resort to standard segmentation metrics in the medical segmentation literature, which includes the DICE coefficient (DSC) and the Average Surface Distance (ASD). To evaluate the calibration performance, we employ both the expected calibration error (ECE) \citep{naeini2015obtaining} and classwise expected calibration error (CECE). The reason to include CECE is because ECE only considers the softmax probability of the predicted class, ignoring the other scores in the softmax distribution \citep{mukhoti2020calibrating}. To compute the ECE given a finite number of samples, we group predictions into $M$ equispaced bins. Let $B_{i}$ denote the set of samples with confidences belonging to the $i^{th}$ bin. The accuracy $A_{i}$ of this bin is computed as $A_{i}=\frac{1}{|B_{i}|}\sum_{j \in B_{i}}1(\hat{y_{j}} =y_{j})$, where 1 is the indicator function, and $\hat{y_{j}}$ and $y_{j}$ are the predicted and ground-truth labels for the $j^{th}$ sample. Similarly, the confidence $C_{i}$ of the $i^{th}$ bin is computed as $C_{i}=\frac{1}{|B_{i}|}\sum_{j \in B_{i}}\hat{p}_{j}$, i.e. $C_{i}$ is the average confidence of all samples in the bin. The ECE can be approximated as a weighted average of the absolute difference between the accuracy and confidence of each bin: }
\rev{

\begin{align}
\label{eq:ece}
ECE = \sum_{i=1}^{M}\frac{|B_{i}|}{N}|A_{i} - C_{i}|
\end{align}

The ECE metric only considers the probability of the predicted class, without considering the other scores in the softmax distribution. A stronger definition of calibration would require the probabilities of all the classes in the softmax distrubution to be calibrated. 
This can be achieved with a simple classwise extension of the ECE metric: Classwise ECE, given by

\begin{align}
\label{eq:cece}
CECE = \sum_{i=1}^{M}\sum_{j=1}^{K}\frac{|B_{i,j}|}{N}|A_{i,j} - C_{i,j}|
\end{align}

where $K$ is the number of classes, $B_{ij}$ denotes the set of samples from the $j^{th}$ class in the $i^{th}$ bin, $A_{i,j}=\frac{1}{|B_{i,j}|}\sum_{k \in B_{i,j}}1(j =y_{k})$ and $C_{i,j}=\frac{1}{|B_{ij}|}\sum_{k \in B_{i,j}}\hat{p}_{kj})$ }

\rev{Following the recent literature on calibration of segmentation networks \citep{islam2021spatially} both ECE and CECE are obtained by considering only the foreground regions. The reason behind this is that most of the correct --and certain-- predictions are from the background. If we exclude these areas from the statistics, the obtained results will better highlight the differences among the different approaches. In our implementation, the number of bins to compute ECE and CECE is set to $M=15$. Furthermore, we also employ reliability plots \citep{niculescu2005predicting} in our evaluation, which plot the expected accuracy as a function of class probability (confidence), and for a perfectly calibrated model it represents the identity function.}

\begin{table*}[h!]
\tiny
\centering
\caption{\rev{The discriminative performance (DSC and ASD) obtained by the different models across seven popular medical image segmentation benchmarks. Best method is highlighted in bold, whereas second best approach is underlined.}}
\label{tab:main-disc}
\begin{tabular}{c|c|cccccccccccccc}
\toprule
Dataset &  Region & \multicolumn{2}{c}{CE} & \multicolumn{2}{c}{CE + DICE} & \multicolumn{2}{c}{FL} & \multicolumn{2}{c}{ECP} & \multicolumn{2}{c}{LS} & \multicolumn{2}{c}{SVLS} & \multicolumn{2}{c}{Ours} \\ 
\midrule
& & DSC \revf{$\uparrow$} & ASD \revf{$\downarrow$} & DSC \revf{$\uparrow$} & ASD \revf{$\downarrow$} & DSC \revf{$\uparrow$} & ASD \revf{$\downarrow$} & DSC \revf{$\uparrow$} & ASD \revf{$\downarrow$} & DSC \revf{$\uparrow$} & ASD \revf{$\downarrow$} & DSC \revf{$\uparrow$} & ASD \revf{$\downarrow$} & DSC \revf{$\uparrow$} & ASD \revf{$\downarrow$}\\
\midrule
\multirow{4}{*}{ACDC} & RV & 0.813 & 0.77 & 0.798 & 0.75 & 0.714 & 1.27 & 0.754 & 0.87 & \und{0.815} & \und{0.68} & 0.642 & 1.86 & \bf 0.866 & \bf 0.42 \\ 
 & MYO & \und{0.816} & 0.43 & 0.795 & 0.46 & 0.734 & 0.61 & 0.751 & 0.53 & 0.805 & \und{0.42} & 0.664 & 1.79 &  \bf 0.845 & \bf 0.37 \\ 
 & LV & \und{0.894} & 0.36 & 0.888 & 0.35 & 0.846 & 0.47 & 0.839 & 0.41 & 0.886 & \und{0.32} & 0.795 & 1.21 &  \bf 0.913 & \bf 0.29 \\ 
 \rowcolor{gray!20} & Mean & \und{0.841} & 0.52 & 0.827 & 0.52 & 0.764 & 0.78 & 0.782 & 0.60 & 0.835 & \und{0.48} & 0.701 & 1.62 & \bf 0.875 & \bf 0.36 \\  
 \midrule
\multirow{4}{*}{MRBrainS} & GM & 0.780 & \und{0.42} & 0.757 & 0.48 & 0.773 & 0.53 & \und{0.793} & 0.47 & 0.745 & 0.51 & 0.753 & 0.49 & \bf 0.800 & \bf 0.39 \\ 
 & WM & \und{0.811} & 0.62 & 0.761 & 0.66 & 0.804 & 0.60 & 0.810 & \und{0.55} & 0.727 & 0.97 & 0.670 & 1.06 & \bf 0.831 & \bf 0.46 \\ 
 & CSF & 0.772 & 0.44 & 0.780 & 0.46 & 0.793 & 0.40 & 0.803 & \und{0.39} & 0.772 & 0.46 & \bf 0.810 & 0.39 & \und{0.807} & \bf 0.38 \\ 
  \rowcolor{gray!20} & Mean & 0.788 & 0.50 & 0.766 & 0.54 & 0.790 & 0.51 & \und{0.802} & \und{0.47} & 0.748 & 0.64 & 0.744 & 0.65 & \bf 0.813 & \bf 0.41 \\  
  \midrule
 \multirow{5}{*}{FLARE} & Liver & 0.949 & 0.60 & 0.942 & \und{0.43} & 0.951 & \bf 0.37 & \und{0.952} & 0.56 & 0.952 & 1.44 & 0.949 & 1.47 & \bf 0.953 & 1.52 \\
 & Kidney & 0.944 & 0.37 & 0.941 & 0.37 & 0.946 & \und {0.32} & \bf 0.950 & \bf 0.31 & \und{0.947} & 0.38 & 0.946 & 0.40 & 0.945 & 0.35 \\
 & Spleen & 0.929 & 0.56 & 0.904 & 0.61 & 0.924 & 0.55 & 0.924 & 0.68 & \bf 0.942 & \und{0.38} & 0.932 & 0.56 & \und{0.940} & \bf 0.38 \\
 & Pancreas & 0.635 & 1.55 & 0.634 & \bf 1.41 & 0.625 & 1.65 & \bf 0.649 & 1.47 & 0.636 & 1.56 & 0.636 & 1.53 & 
 \und{0.645} & \und{1.42} \\
 \rowcolor{gray!20} & Mean & 0.864 & 0.77 & 0.855 & \bf 0.71 & 0.862 & \und {0.72} & \und{0.869} & 0.75 & 0.869 & 0.94 & 0.866 & 0.99 & \bf 0.871 & 0.92 \\
 \midrule
\multirow{4}{*}{BRATS} & TC & \revf{0.832} & \revf{\und {2.36}} & \revf{0.746} & \revf{4.98} & \revf{\und{0.854}} & \revf{3.13} & \revf{0.834} & \revf{2.63} & \revf{0.807} & \revf{2.96} & \revf{0.763} & \revf{3.48} & \bf \revf{0.856} & \bf \revf{2.24}  \\ 
 & ET & \revf{0.794} & \bf \revf{1.59} & \revf{0.729} & \revf{3.23} & \revf{\und{0.799}} & \revf{2.58} & \revf{0.783} & \revf{1.81} & \revf{0.773} & \revf{\und{1.59}} & \revf{0.749} & \revf{2.31} & \bf \revf{0.811} & \revf{1.62} \\
& WT & \revf{\und{0.889}} & \revf{2.66} & \revf{0.854} & \revf{2.93} & \revf{0.889} & \revf{2.72}  & \revf{0.889} & \revf{2.41} & \revf{0.879} & \revf{2.48} & \revf{0.888} & \revf{\und{2.27}} & \bf \revf{0.895} & \bf \revf{2.11} \\ 
\rowcolor{gray!20} & Mean & \revf{0.839} & \revf{\und{2.20}} & \revf{0.776} & \revf{3.71} & \revf{\und{0.848}} & \revf{2.81} & \revf{0.836} & \revf{2.28} & \revf{0.819} & \revf{2.34} & \revf{0.798} & \revf{2.69} & \bf \revf{0.854} & \bf \revf{1.99} \\  
  \midrule
\multirow{3}{*}{PROMISE}  & Prostate & 0.737 & 1.33 & 0.751 & \und{1.17} & 0.729 & 1.42 & 0.736 & 1.27 & 0.713 & 1.72 & \und{0.766} & 1.27 & \bf 0.770 & \bf 0.95 \\  
 & Tumor & 0.258 & 5.81 & 0.328 & 4.10 & 0.361 & 3.35 & 0.344 & 2.48 & 0.350 & 3.29 & \und{0.396} & \bf 2.16 & \bf 0.397 & \und{2.34} \\ 
 \rowcolor{gray!20} & Mean & 0.498 & 3.57 & 0.540 & 2.63 & 0.545 & 2.39 & 0.540 & 1.88 & 0.532 & 2.50 & \und{0.581} & \und{1.71} & \bf 0.583 & \bf 1.64 \\ 
  \midrule  
\multirow{3}{*}{\revf{HPC}} & \revf{Anterior} & \revf{0.876} & \revf{0.51} & \revf{0.8741} & \revf{0.47} & \revf{\und{0.879}} & \revf{\und{0.46}} & \revf{0.874} & \revf{0.49} & \revf{0.879} & \revf{0.49} & \revf{\bf 0.883} & \revf{\bf {0.46}} & \revf{0.876} & \revf{0.49} \\  
 & \revf{Posterior} & \revf{\bf {0.858}} & \revf{0.43} & \revf{0.857} & \revf{0.43} & \revf{0.852} & \revf{0.48} & \revf{0.853} & \revf{0.45} & \revf{0.857} & \revf{\bf{0.42}} & \revf{0.849} & \revf{0.48} & \revf{\und{0.857}} & \revf{\und{0.43}} \\ 
 \rowcolor{gray!20} & \revf{Mean} & \revf{\und{0.867}} & \revf{0.47} & \revf{0.865} & \revf{\bf{0.45}} & \revf{0.865} & \revf{0.47} & \revf{0.864} & \revf{0.47} & \revf{\bf{0.868}} & \revf{\und{0.45}} & \revf{0.866} & \revf{0.47} & \revf{\und{0.867}} & \revf{0.46} \\ 
  \midrule  
\revf{BUSI} & \revf{Tumor} & \revf{0.673} & \revf{\und{13.2}} & \revf{\bf 0.709} & \revf{\bf 13.1} & \revf{\und{0.688}} & \revf{13.9} & \revf{0.677} & \revf{15.1} & \revf{0.679} & \revf{15.6} & \revf{0.679} & \revf{14.6} & \revf{0.685} & \revf{13.7} \\  
\bottomrule
\end{tabular}
\end{table*}

\begin{table*}[h!]
\tiny
\centering
\caption{\rev{The calibration performance (ECE and CECE) obtained by the different models across five popular medical image segmentation benchmarks. Best method is highlighted in bold, whereas second best approach is underlined. $\nabla$ indicates the difference between the best model and our approach. }}
\label{tab:main-cal}
\begin{tabular}{c|cccccccccccccccc}
\toprule
Dataset &   \multicolumn{2}{c}{CE} & \multicolumn{2}{c}{CE + DICE} & \multicolumn{2}{c}{FL} & \multicolumn{2}{c}{ECP} & \multicolumn{2}{c}{LS} & \multicolumn{2}{c}{SVLS} & \multicolumn{2}{c}{Ours} \\ 
\midrule
& ECE \revf{$\downarrow$} & CECE \revf{$\downarrow$} & ECE \revf{$\downarrow$} & CECE \revf{$\downarrow$} & ECE \revf{$\downarrow$} & CECE \revf{$\downarrow$} & ECE \revf{$\downarrow$} & CECE \revf{$\downarrow$} & ECE \revf{$\downarrow$} & CECE \revf{$\downarrow$} & ECE \revf{$\downarrow$} & CECE \revf{$\downarrow$} & ECE \revf{$\downarrow$} & CECE \revf{$\downarrow$} & $\nabla$ECE & $\nabla$CECE  \\
\midrule
ACDC  & \und{0.079} & \und{0.073} & 0.137 & 0.084 & 0.113 & 0.116 & 0.109 & 0.095 & 0.081 & 0.107 & 0.176 & 0.135 & \bf{0.061} & \bf 0.069 & -- & --\\ 
MRBrainS & 0.089 & 0.070 & 0.172 & 0.102 & \bf 0.020 & \und{0.064} & 0.048 & 0.068 & \und{0.036} & 0.085 & 0.060 & 0.080 & 0.050 & \bf 0.058 & 0.030 & -- \\ 
FLARE & 0.045 &  0.029 & 0.058 &  0.034 & \bf 0.033 & 0.035 & \und{0.037} & \bf 0.027 & 0.055 & 0.050 & 0.039 & 0.036 & 0.038 & \und{0.028} & 0.005 & 0.001 \\ 
\revf{BRATS} & \revf{0.131} & \revf{\bf 0.091} & \revf{0.178} & \revf{0.122} & \revf{\bf 0.097} & \revf{0.119} & \revf{0.132} & \revf{\und {0.091}} & \revf{0.112} & \revf{0.108} & \revf{0.151} & \revf{0.122} & \revf{\und{0.101}} & \revf{0.093} & \revf{0.004} & \revf{0.002} \\ 
PROMISE  & 0.411 & 0.334 & 0.430 & 0.304 & \und{0.247} & 0.298 & 0.306 & \und{0.252} & 0.280 & 0.299 & 0.344 & 0.271 & \bf 0.232 & \bf 0.237 & -- & -- \\ 
\revf{HPC}  & \revf{0.052} & \revf{0.091} & \revf{0.069} & \revf{\bf{0.079}} & \revf{\und{0.042}} & \revf{0.108} & \revf{0.066} & \revf{0.093} & \revf{0.061} & \revf{0.109} & \revf{0.044} & \revf{0.104} & \revf{\bf 0.033} & \revf{\und{0.088}} & \revf{--} & \revf{0.009} \\ 
\revf{BUSI} & \revf{0.230} & \revf{0.334} & \revf{0.250} & \revf{\und{0.278}} & \revf{\und{0.220}} & \revf{0.305} & \revf{0.237} & \revf{0.365} & \revf{0.229} & \revf{0.333} & \revf{0.226} & \revf{0.305} & \revf{\bf 0.193} & \revf{\bf 0.274} & \revf{--} & \revf{--} \\ 
\bottomrule
\end{tabular}
\end{table*}

\subsubsection{\rev{Implementation Details}}

\rev{To empirically evaluate the proposed model, we conduct experiments comparing a state-of-the-art segmentation network on a multi-class scenario trained with different learning objectives. In particular, we first include standard loss functions employed in medical image segmentation, which include the common Cross-entropy (CE) and the duple composed by CE and DSC losses. Furthermore, we also include training objectives which have been proposed to calibrate neural networks, which represent nowadays the state-of-the-art for this task. This includes Focal loss (FL) \citep{lin2017focal}, Label Smoothing (LS) \citep{szegedy2016rethinking} and ECP  \citep{pereyra2017regularizing}. Last, we also compare to the recent Spatially-Varying LS (SVLS) \citep{islam2021spatially}, which demonstrated to outperform the simpler LS version in the task of medical image segmentation. Following the literature, we have chosen the commonly used hyper-parameters and considered the values which provided \revf{the best DSC on the validation dataset}. For FL, $\gamma$ values of 1, 2, and 3 are considered. In case of ECP and LS, $\alpha$ and  $\lambda$ values of 0.1, 0.2, 0.3 are used. For our method, we considered the margins to be 5, 8, and 10. In the case of SVLS, the one-hot label smoothing is performed with a kernel size of 3. For the experiments, we fixed the batch size to 4, epochs to 100, and optimizer to ADAM. The learning rate of 1e-3 and 1e-4 are used for the first 50 epochs, and the next 50 epochs respectively.} \revf{The models are trained on 2D slices, while the evaluation is done over 3D volumes.}

\noindent \rev{\textbf{Backbones.} The main experiments are conducted on the popular UNet \citep{unet}. Nevertheless, to show the versatility of the proposed margin based label smoothing, we have evaluated our model on other popular architectures in medical image segmentation including AttUNet \citep{attunet}, TransUNet \citep{transunet}, and UNet++ \citep{unetpp}.}

\subsection{\rev{Results}}

\subsubsection{\rev{Main results}}

\rev{The discriminative quantitative results obtained by the proposed model, as well as prior literature, are reported in Table \ref{tab:main-disc}. We observe that across the different datasets, our model consistently achieves \revf{competitive} discriminative performance, typically ranking as the best or second-best model in both region-based (i.e., DSC) and distance-based (i.e., ASD) metrics. 
This demonstrates that our method yields not only a better identification of target regions, but also an improvement in the boundary regions, highlighted by lower ASD values. An interesting observation is that, while other learning objectives typically result in performance gains compared to the standard CE loss, their superiority over the others depends on the selected dataset.}

\begin{figure*}[h!]
    \begin{center}
    \includegraphics[width=.95\linewidth]{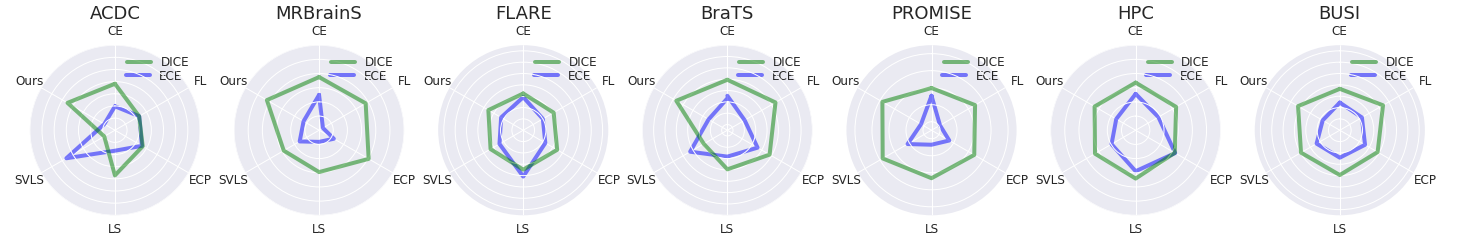}
    \end{center}
    \vspace{-1 em}
    \caption{\rev{\textbf{Compromise between calibration and discriminative performance.} For each dataset, we show the discriminative (DICE) and calibration (ECE) results for each method. In order to get the best performance, we expect a model to achieve large DSC (\textit{in green}) and small ECE (\textit{in blue}) values.}}
     \label{fig:main}
\end{figure*}

\rev{Table \ref{tab:main-cal} summarizes the calibration performance, in terms of ECE and CECE of all the analyzed models. We can observe that, similar to the discriminative performance reported earlier, the proposed model typically ranks as best or second best method. 
An interesting observation is that, according to the results, focal loss provides well-calibrated models \revf{(i.e., low ECE and CECE values)}, whereas their discriminative performance is typically far from best performing models. As exposed in our motivation, one of the reasons behind this behaviour might be the undesirable effect of pushing all logit distances to zero. Enforcing this constraint may alleviate the problem of overconfidence in deep networks, at the cost of providing non-informative solutions. }

\rev{An interesting summary of these results is depicted in Figure \ref{fig:main}, where we resort to radar plots to highlight the better compromise between discriminative and calibration performance shown by our model. In particular, a \textit{well-calibrated} model should have a balanced compromise between a high discriminative power (\textit{green line}) and low calibration metrics (\textit{\revf{blue} line}). This means that, following these plots, the larger the gap between green and \revf{blue} lines, the better the compromise between discriminative and calibration performance.}

\begin{table*}[t]
\tiny
\centering
\caption{\rev{Calibration performance of post-hoc calibration methods: temperature scaling (TS) and Local Temperature Scaling (LTS) \citep{ding2021local}. Best method is highlighted in bold, whereas second best approach is underlined.}}
\label{tab:temp_scale}
\begin{tabular}{c|c|cccccccccccccc}
\toprule
Dataset & Method & \multicolumn{2}{c}{CE} & \multicolumn{2}{c}{CE + DICE} & \multicolumn{2}{c}{FL} & \multicolumn{2}{c}{ECP} & \multicolumn{2}{c}{LS} & \multicolumn{2}{c}{SVLS} & \multicolumn{2}{c}{Ours} \\ 
\midrule
& & ECE & CECE & ECE & CECE & ECE & CECE & ECE & CECE & ECE & CECE & ECE & CECE & ECE & CECE\\
\midrule
\multirow{4}{*}{ACDC} & Pre & \und{0.079} & \und{0.073} & 0.137 & 0.084 & 0.113 & 0.116 & 0.109 & 0.095 & 0.081 & 0.107 & 0.176 & 0.135 & \bf 0.061 & \bf 0.069 \\ 
 & TS & \und{0.077} & \und{0.073} & 0.135 & 0.084 & 0.112 & 0.117 & 0.105 & 0.095 & 0.084 & 0.109 & 0.174 & 0.135 &  \bf 0.055 & \bf 0.065 \\ 
 & LTS & 0.067 & \und{0.065} & 0.070 & 0.076 & 0.127 & 0.125 & 0.080 & 0.094 & \und {0.065} & 0.072 & 0.118 & 0.116 &  \bf 0.041 & \bf 0.046 \\ 
 \midrule
\multirow{4}{*}{FLARE} & Pre & 0.045 & 0.029 & 0.058 & 0.034 & \bf 0.033 & 0.035 & \und{0.037} & \bf 0.027 & 0.055 & 0.050 & 0.039 & 0.036 & 0.038 & \und{0.028} \\ 
 & TS & 0.040 & 0.030 & 0.051 & 0.036 & \bf 0.030 & 0.038 & \und{0.032} & \bf 0.028 & 0.042 & 0.039 & 0.039 & 0.038 & 0.033 & \und {0.029} \\ 
 & LTS & 0.033 & 0.030 & 0.044 & 0.038 & 0.065 & 0.048 & \bf 0.026 & 0.028 & 0.031 & \und{0.026} & 0.040 & 0.036 &  \und{0.031} & \bf 0.026 \\ 
 \midrule
 \multirow{4}{*}{\revf{BRATS}} & \revf{Pre} & \revf{0.131} & \revf{\bf{0.091}} & \revf{0.178} & \revf{0.122} & \revf{\bf{0.097}} & \revf{0.119} & \revf{0.132} & \revf{\und{0.091}} & \revf{0.112} & \revf{0.108} & \revf{0.151} & \revf{0.122} & \revf{\und{0.101}} & \revf{0.093} \\ 
 & \revf{TS} & \revf{0.13} & \revf{\bf{0.09}} & \revf{0.177} & \revf{0.122} & \revf{\bf{0.097}} & \revf{0.119} & \revf{0.131} & \revf{\und{0.091}} & \revf{0.111} & \revf{0.108} & \revf{0.149} & \revf{0.121} & \revf{\und{0.098}} & \revf{0.093} \\ 
 & \revf{LTS} & \revf{0.114} & \revf{\bf{0.089}} & \revf{0.156} & \revf{0.121} & \revf{\und{0.097}} & \revf{0.119} & \revf{0.117} & \revf{\und{0.09}} & \revf{0.105} & \revf{0.119} & \revf{0.131} & \revf{0.121} & \revf{\bf{0.089}} & \revf{0.096} \\ 
 \midrule
\multirow{4}{*}{PROMISE} & Pre & 0.411 & 0.334 & 0.430 & 0.304 & \und{0.247} & 0.298 & 0.306 & \und{0.252} & 0.280 & 0.299 & 0.344 & 0.271 & \bf 0.232 & \bf 0.237 \\ 
 & TS & 0.408 & 0.334 & 0.429 & 0.304 & \und{0.245} & 0.299 & 0.303 & \und{0.251} & 0.279 & 0.298 & 0.342 & 0.271 & \bf 0.229 & \bf 0.237 \\ 
 & LTS & 0.294 & 0.283 & 0.312 & 0.263 & \und{0.209} & 0.291 & 0.230 & \und{0.235} & 0.255 & 0.257 & 0.234 & 0.238 & \bf 0.189 & \bf 0.217 \\ 
   \midrule
\multirow{4}{*}{\revf{HPC}} & \revf{Pre} & \revf{0.052} & \revf{0.091} & \revf{0.069} & \revf{\bf{0.079}} & \revf{\und{0.042}} & \revf{0.108} & \revf{0.066} & \revf{0.093} & \revf{0.061} & \revf{0.109} & \revf{0.044} & \revf{0.104} & \revf{\bf 0.034} & \revf{\und{0.088}} \\ 
 & \revf{TS} & \revf{0.051} & \revf{0.091} & \revf{0.068} & \revf{\bf{0.079}} & \revf{\und{0.042}} & \revf{0.108} & \revf{0.065} & \revf{0.093} & \revf{0.059} & \revf{0.108} & \revf{0.044} & \revf{0.104} & \revf{\bf{0.033}} & \revf{\und{0.089}} \\ 
 & \revf{LTS} & \revf{0.048} & \revf{0.092} & \revf{0.065} & \revf{\bf{0.08}} & \revf{\und{0.041}} & \revf{0.108} & \revf{0.061} & \revf{0.094} & \revf{0.059} & \revf{0.108} & \revf{0.043} & \revf{0.103} & \revf{\bf{0.032}} & \revf{\und{0.09}} \\ 
 \midrule
\multirow{4}{*}{\revf{BUSI}} & \revf{Pre} & \revf{0.230} & \revf{0.334} & \revf{0.250} & \revf{\und{0.278}} & \revf{\und{0.220}} & \revf{0.305} & \revf{0.237} & \revf{0.365} & \revf{0.229} & \revf{0.333} & \revf{0.226} & \revf{0.305} & \revf{\bf 0.193} & \revf{\bf 0.274} \\ 
 & \revf{TS} & \revf{0.229} & \revf{0.333} & \revf{0.250} & \revf{\und{0.278}} & \revf{0.236} & \revf{0.365} & \revf{\und{0.219}} & \revf{0.305} & \revf{0.229} & \revf{0.333} & \revf{0.225} & \revf{0.305} & \revf{\bf 0.193} & \revf{\bf 0.274} \\ 
 & \revf{LTS} & \revf{0.207} & \revf{0.328} & \revf{0.210} & \revf{\bf{0.257}} & \revf{0.243} & \revf{0.377} &  \revf{0.202} & \revf{0.298} & \revf{0.268} & \revf{0.358} & \revf{\und{0.198}} & \revf{0.295} & \revf{\bf{0.182}} & \revf{\und{0.275}}\\ 
\bottomrule
\end{tabular}
\end{table*}

\begin{figure}[h]
     \centering
     \begin{subfigure}[b]{0.49\linewidth}
         \centering
         \includegraphics[width=\linewidth]{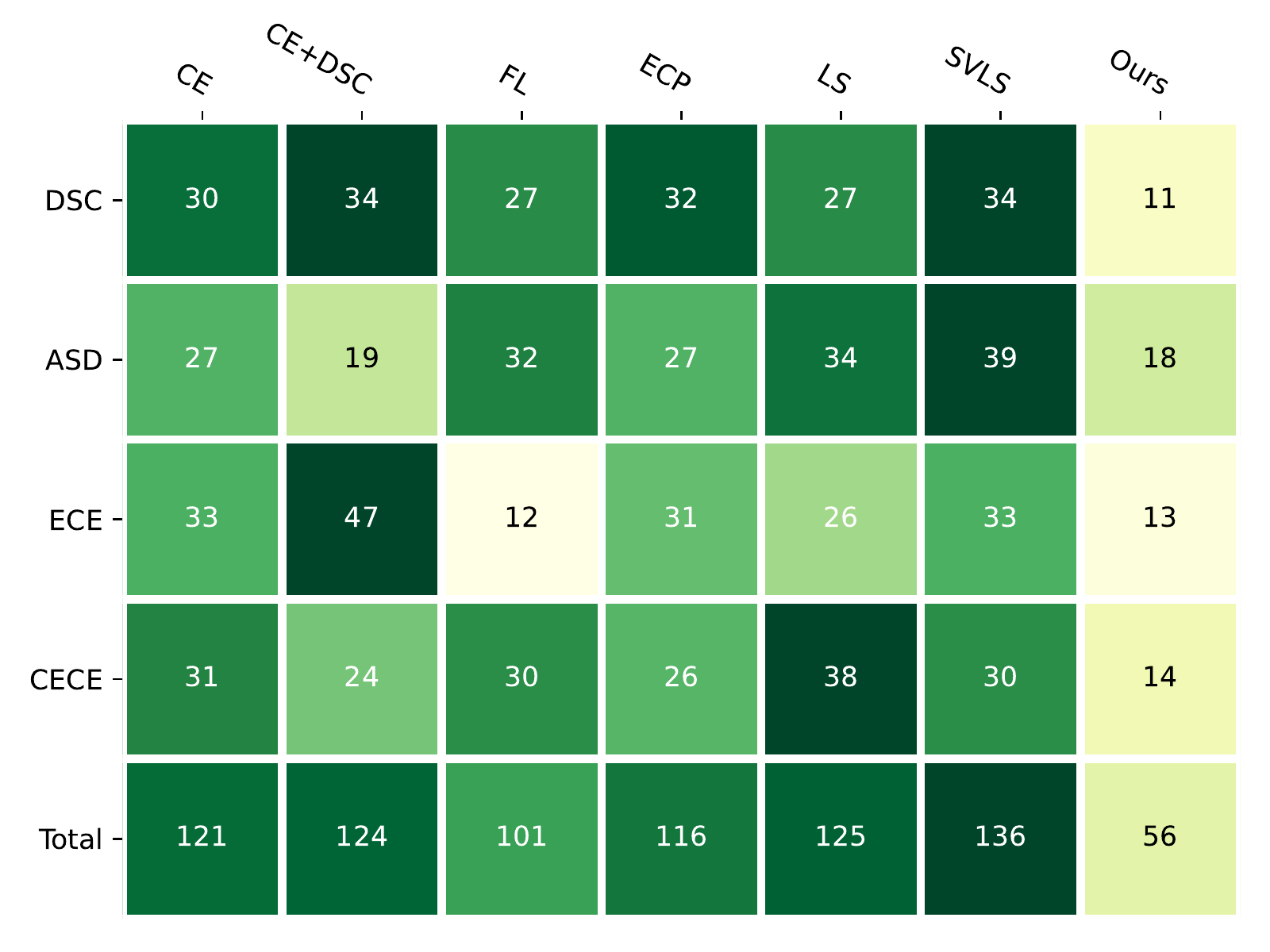}
         \caption{}
         \label{fig:sum-rank}
     \end{subfigure}
     \begin{subfigure}[b]{0.49\linewidth}
         \centering
         \includegraphics[width=\linewidth]{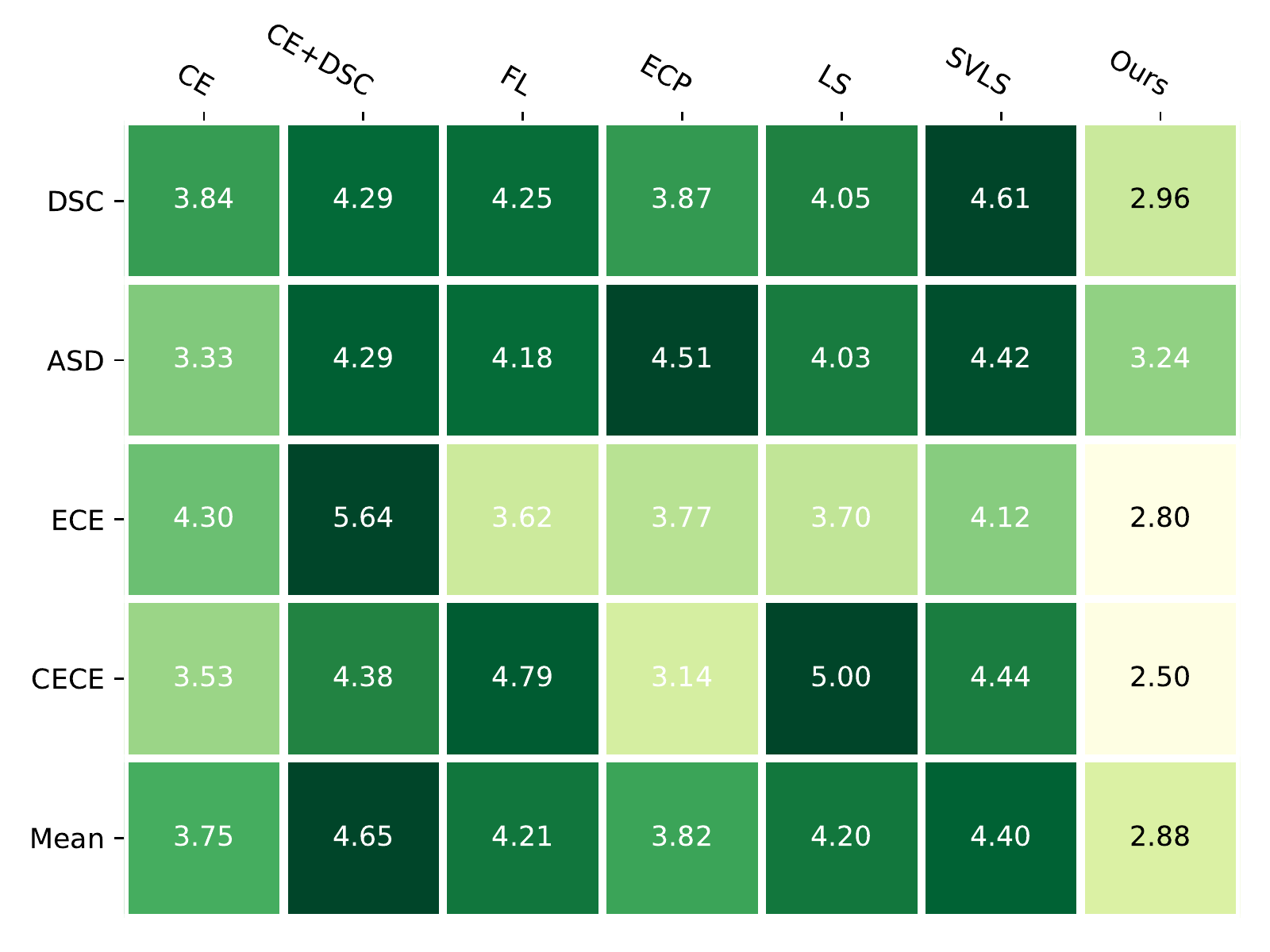}
         \caption{}
         \label{fig:mean-case-rank}
     \end{subfigure}
    \caption{\revf{Ranking (\textit{global} and \textit{per-metric}) of the different methods based on the sum-rank and mean of case-specific approach.}}
\end{figure}

\rev{Furthermore, to have a better overview of the general performance across different models, we follow the strategy followed in several MICCAI Challenges, e.g., MRBrainS \citep{mendrik2015mrbrains}, where the final ranking is given as the sum of individual ranking metrics: 
$\rev{ R_{T} = \sum_{m=0}^{|M|}  r_{m}}$, where $r_{m}$
is the rank of the segmentation model for the metric $m$ (mean)\footnote{\rev{Note that the per-class scores are not used in the sum-rank computation.}}. Thus, if a model ranks first in terms of DSC in the FLARE dataset, it will receive one point, whereas five points will be added in case the model ranks fifth. The final ranking is obtained after the overall scores $R_{T}$ for each model are sorted in ascending order, and ranked from $1$ to $n$. \revf{Furthermore, to account for the different complexities of each sample, we follow the mean-case-rank strategy, which has been employed in other MICCAI Challenges, e.g., \citep{MAIER2017250}. In particular, we first compute the DSC, ASD, ECE, and CECE values for each sample, and establish each method’s rank based on these metrics, separately for each case. Then, we compute the mean rank over all four evaluation metrics, per case, to obtain the method’s rank for that given sample. Finally, we compute the mean over all case-specific ranks to obtain the method’s final rank.} Figure \ref{fig:sum-rank} provides the rank comparison through heatmap visualization. It can be inferred that, for both discriminative and calibration metrics, our methods achieves the highest rank. Interestingly, the proposed loss term yields very competitive discriminative results, outperforming the popular compounded CE+DSC loss. It is noteworthy to highlight that the optimization goal of these two terms are different. Networks trained with CE tend to achieve a lower average negative log-likelihood over all the pixels, whereas using Dice as loss function should increase the discriminative performance, in terms of Dice. Thus, it is expected that the compounded loss brings the better of both worlds. Nevertheless, we can observe that this is not what happens in practice. On the one hand, the networks trained with CE+DSC loss rank among the best discriminative models (third in DSC and second in ASD). On the other hand, their calibration performance is substantially degraded, ranking last and second-last in ECE and CECE, respectively. These results align with recent findings \citep{mehrtash2020confidence}, which highlight the deficiencies of models trained with the DSC loss to deliver well-calibrated models. While adjusting the balancing hyperparameter could improve the performance on one task, the results on the other task would likely degrade due to the different nature of both learning objectives. Thus, based on these observations, we argue that obtaining a good compromise between calibration and segmentation quality is hardly attainable with the popular CE+DSC loss, and promote our model as a better alternative. Fig. \revf{\ref{fig:mean-case-rank} provides a heatmap visualization to compare the methods for different metrics using mean-case-specific strategy. As observed in sum rank approach, our methods consistently achieves the best rank in both discriminative and calibration metrics. Importantly, calibration methods like FL, and LS achieve promising results with ECE, while severely compromising the DSC.}}

\subsubsection{\rev{Comparison to post-hoc calibration}}

\rev{The proposed approach is orthogonal to post-hoc calibration strategies, which can still be used after training, as long as there exists an independent validation set to find the optimal hyperparameters (for example $T$ in temperature scaling). To demonstrate this, we now report the performance of pre-scaling and post-scaling for ACDC, FLARE, and PROMISE datasets across the different approaches. In particular, we have included two post-hoc calibration strategies. First, we use the standard Temperature scaling approach, referred to as TS, where a single value for the entire image is employed. Furthermore, we also include the Local Temperature Scaling (LTS)  method in \citep{ding2021local}, which was recently proposed in the context of medical image segmentation and provides a temperature value at each image pixel.
For both TS and LTS, the optimal temperature values are found by optimizing the network parameters to decrease the negative log likelihood on an independent validation set. From the quantitative comparison, which can be found in Table \ref{tab:temp_scale}, 
it can be inferred that our method further benefits from scaling the raw softmax probability predictions. Interestingly, the calibration performance obtained by our method prior to temperature scaling still outperforms the results obtained by several other approaches even after applying LTS on their predictions. Another unexpected observation is that, under some settings, the use of temperature scaling (either TS or LTS) deteriorates the calibration performance.  
We argue that this phenomenon could be due to noticeable differences between the validation and testing datasets. As empirically demonstrated in \citep{ovadia2019can}, applying temperature scaling when differences between datasets exist might result in a negative impact. In addition, similar observations were reported in \citep{kock2021confidence}, where the calibration performance of segmentation models on several datasets was degraded after applying temperature scaling}.

\subsubsection{\rev{Effects of logit margin constraints}} 

\begin{figure}[h!]
     \centering
     \begin{subfigure}[b]{0.49\linewidth}
         \centering
         \includegraphics[width=\linewidth]{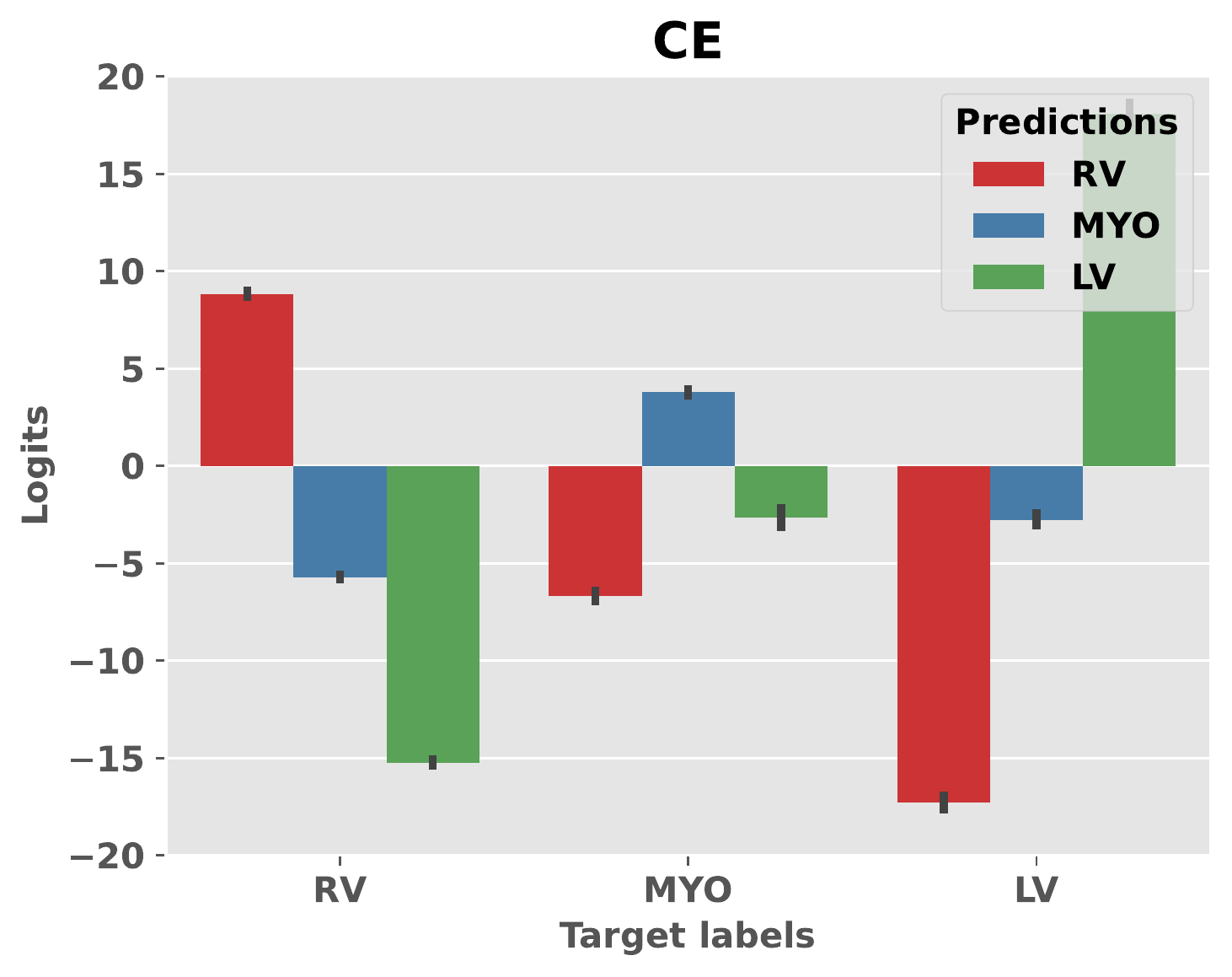}
     \end{subfigure}
     \begin{subfigure}[b]{0.49\linewidth}
         \centering
         \includegraphics[width=\linewidth]{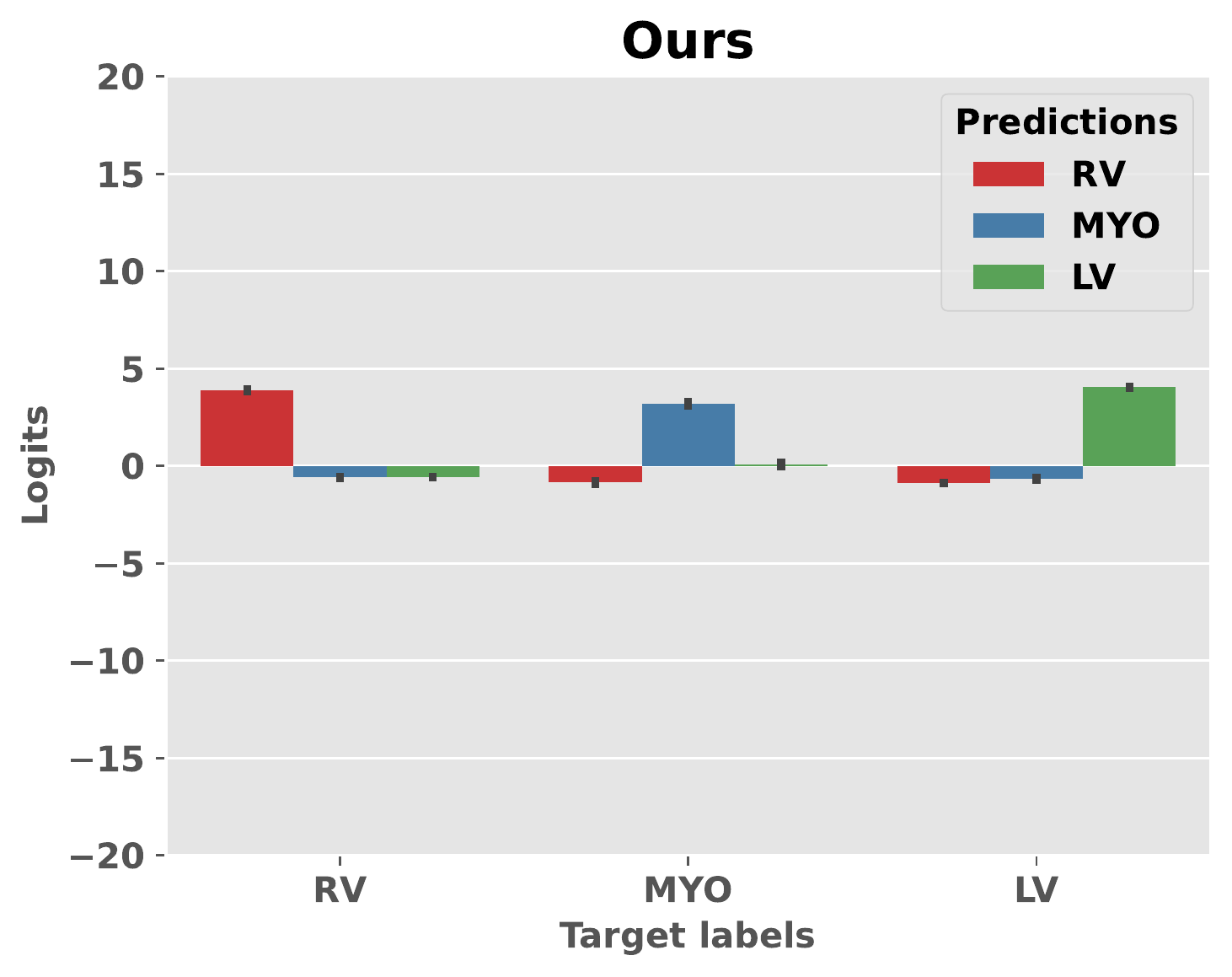}
     \end{subfigure}
     \begin{subfigure}[b]{0.49\linewidth}
         \centering
         \includegraphics[width=\linewidth]{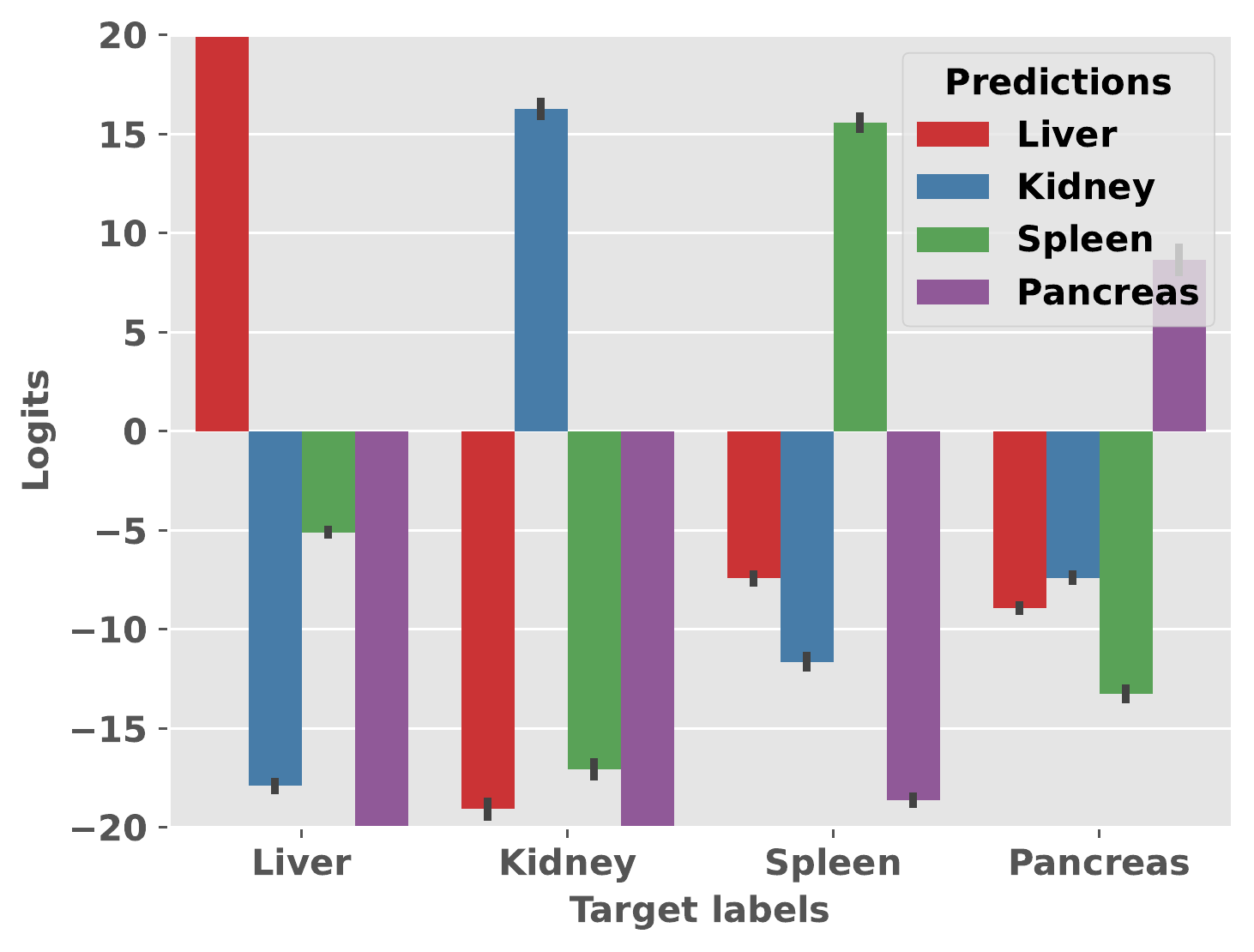}
     \end{subfigure}
     \begin{subfigure}[b]{0.49\linewidth}
         \centering
         \includegraphics[width=\linewidth]{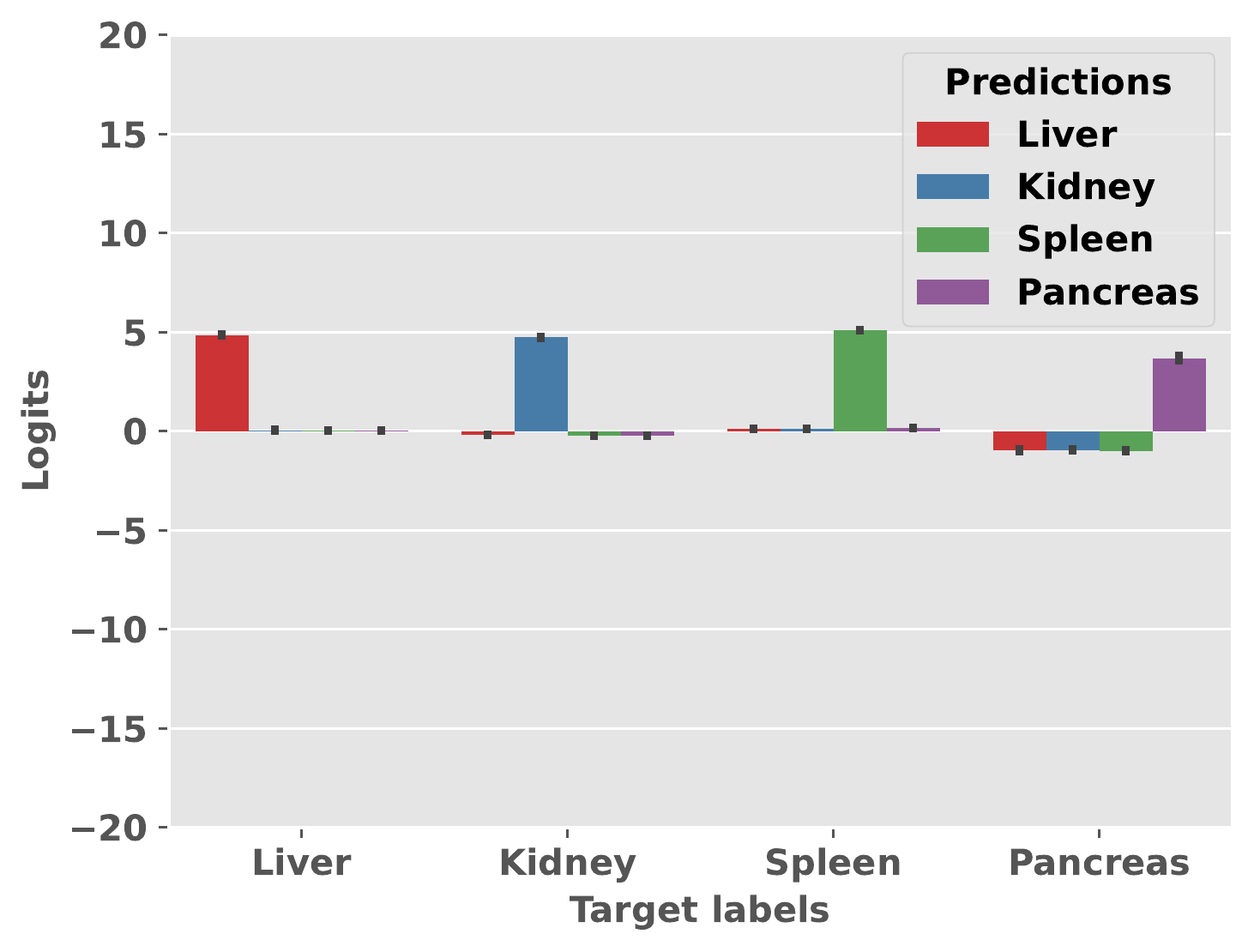}
     \end{subfigure}
    \caption{\rev{\textbf{Adopting the proposed term during training \textit{substantially} reduces the logit distances, producing less overconfident predictions.} These plots depict the average predicted logit distributions for each target class --based on the ground truth-- on ACDC (\textit{top}) and FLARE (\textit{bottom}) datasets when the model is trained with CE (\textit{left}) and the proposed loss (\textit{right}).}}
        \label{fig:logitDist}
\end{figure}

\rev{In our motivation, we hypothesized that the suboptimal supervision delivered by CE in multi-class scenarios might likely result in poorly calibrated models, as the posterior probability assigned to each of the non-true classes cannot be directly controlled. Indeed, it is expected that by minimizing the CE the softmax vectors are pushed towards the vertex of the probability simplex. This implies that the distances between the largest logit and the rest are magnified, resulting in overconfident models. To validate this hypothesis, and to empirically demonstrate that our proposed term can alleviate this issue, we plot the average logit distributions across classes on two datasets. In particular, we first separate all the voxels based on their ground truth labels. Then, for each category group, we average the per-voxel vector of logit predictions for both CE and the proposed model, whose results are depicted in Figure \ref{fig:logitDist}. First, we can observe that a model trained with CE indeed tends to provide large logit differences, which intensifies overconfidence predictions. Furthermore, while the mean logit value of the target class is considerably large and greatly differs from the largest value across other categories, the differences with the remaining logits --from non-target classes-- remain uncontrolled. In contrast, we can clearly observe the impact on the logit distribution when we include the proposed term into the learning objective. In particular, our margin-based term \textit{i)} promotes similar values of the true class logit across classes and \textit{ii)} encourages more equidistant logits between this and the remaining classes, which implicitly constraints the logit values of untargeted classes to be very close (mimicking a uniform distribution). These results empirically validate our hypothesis in regards of the weaknesses of CE and the benefits brought by our approach. }


\begin{figure}[h!]
     \centering
     \begin{subfigure}[b]{0.49\linewidth}
         \centering
         \includegraphics[width=\linewidth]{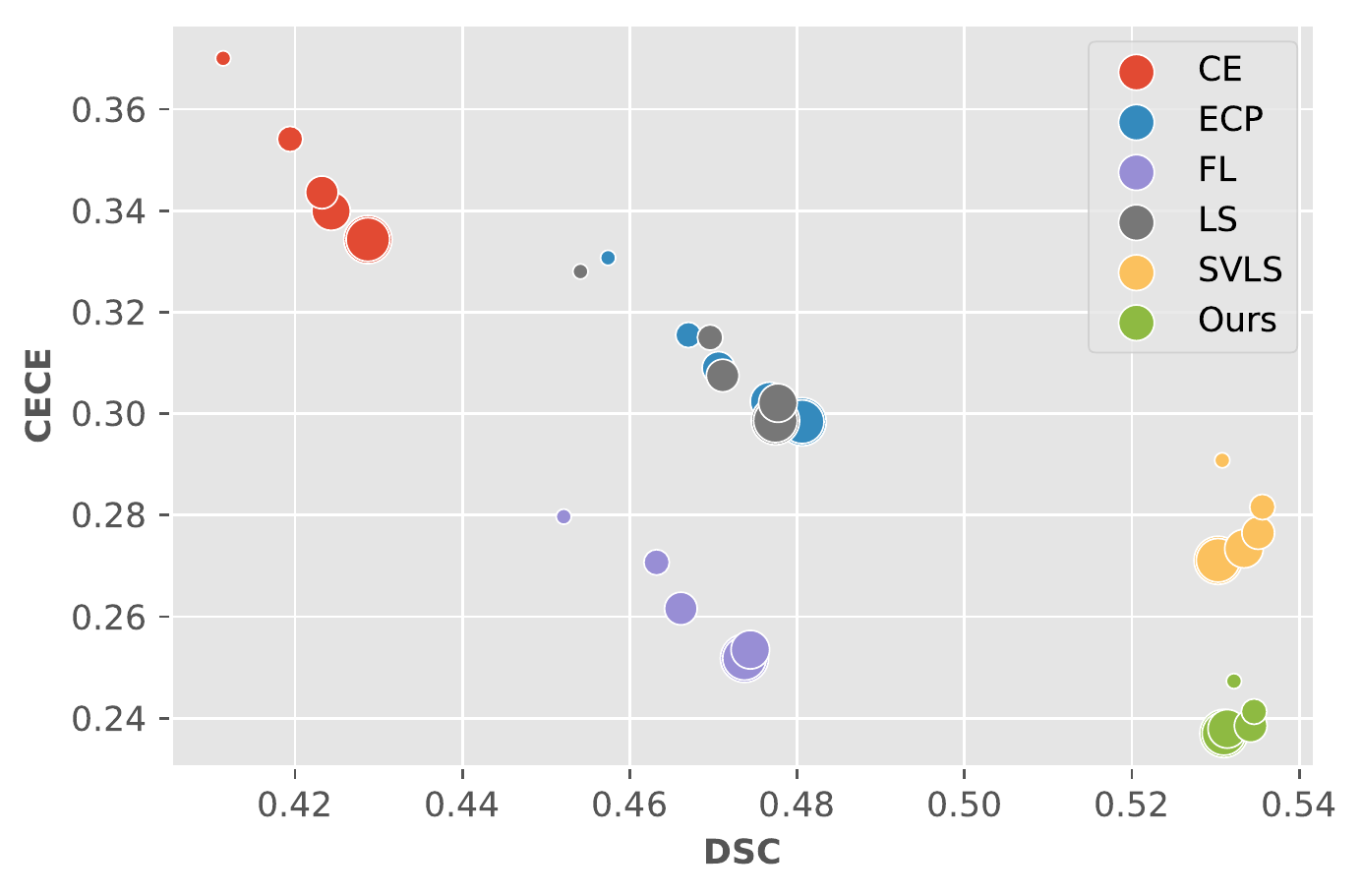}
     \end{subfigure}
     \begin{subfigure}[b]{0.49\linewidth}
         \centering
         \includegraphics[width=\linewidth]{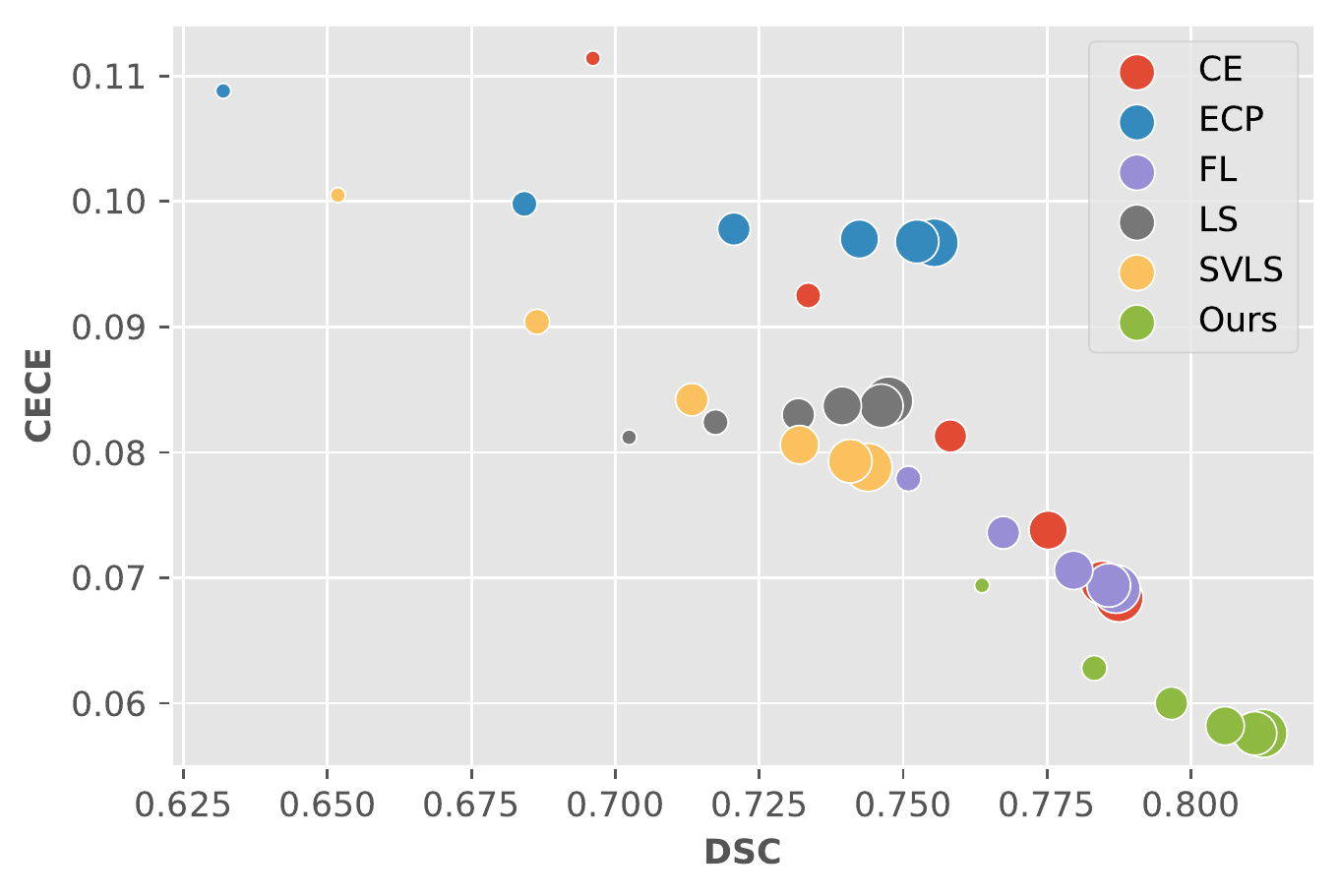}
     \end{subfigure}
        \caption{\rev{Robustness to distributional drift on PROMISE (\textit{left}) and MRBrainS (\textit{right}) datasets. Note that larger circles represent lower sigma values for the Gaussian noise corruptions.}}
        \label{fig:rob_noise}
\end{figure}

\begin{figure*}[h!]
\centering
\begin{subfigure}[b]{.24\linewidth}
\includegraphics[width=\linewidth]{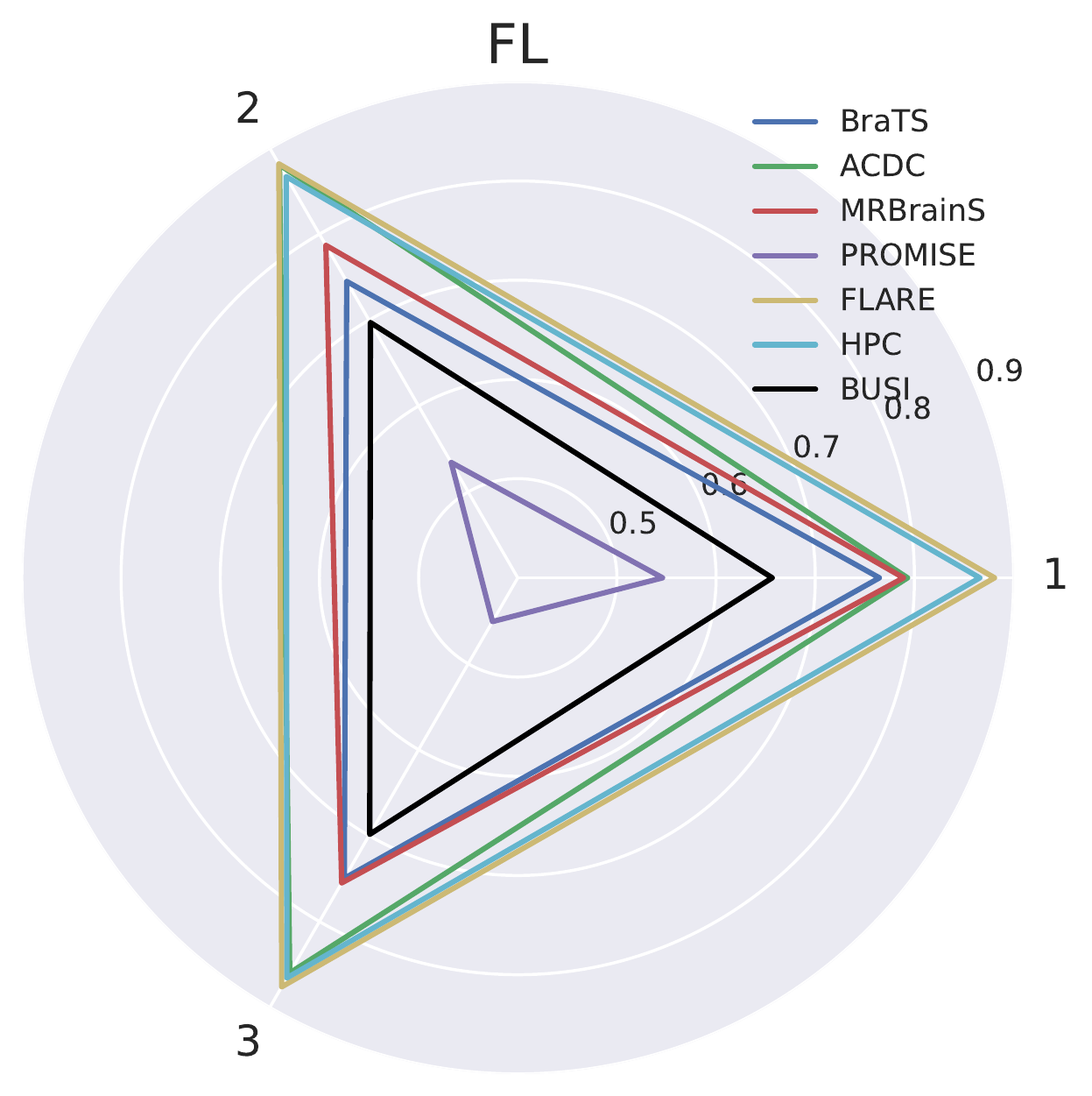}
\end{subfigure}
\begin{subfigure}[b]{.24\linewidth}
\includegraphics[width=\linewidth]{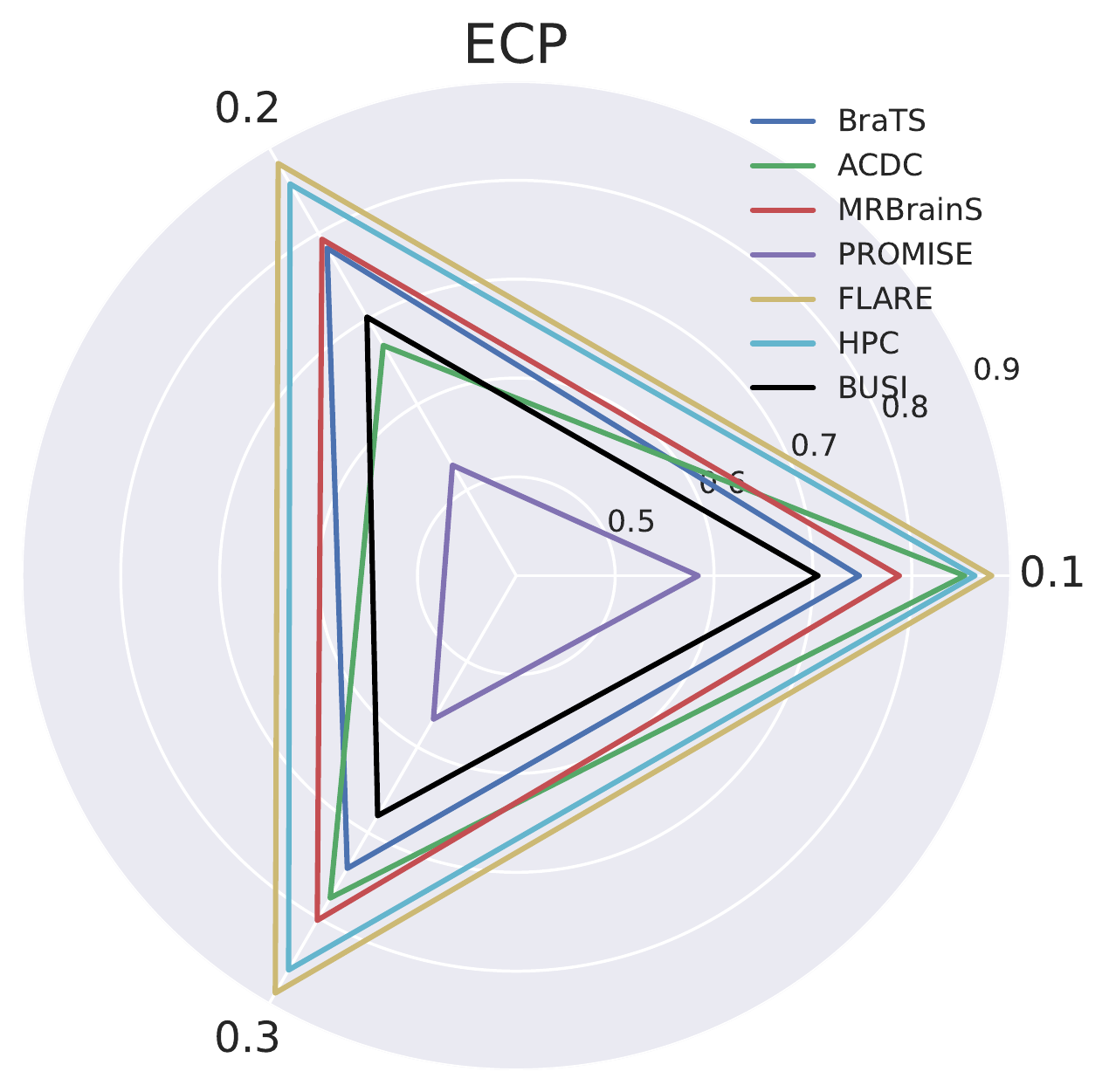}
\end{subfigure}
\begin{subfigure}[b]{.24\linewidth}
\includegraphics[width=\linewidth]{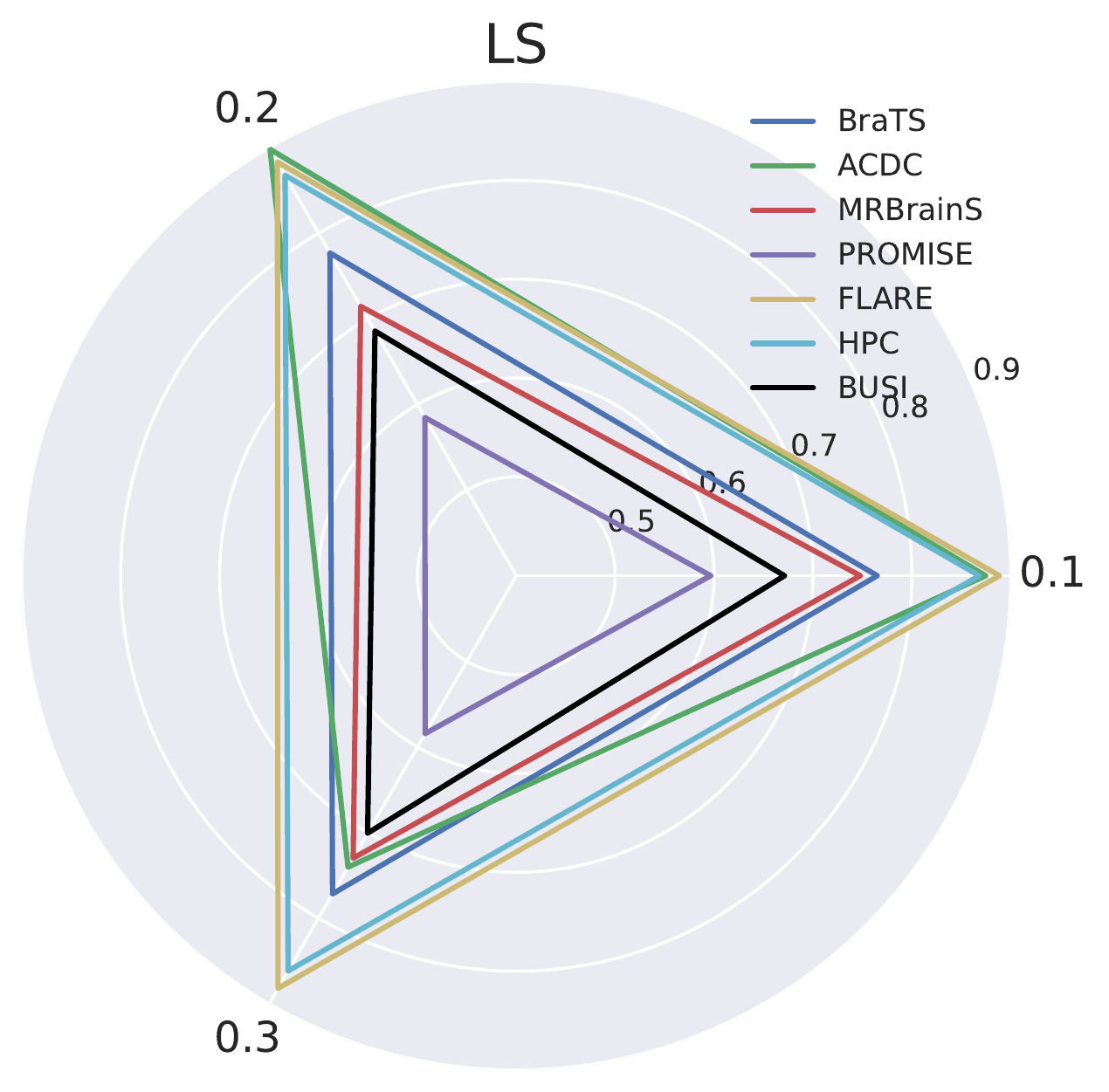}
\end{subfigure}
\begin{subfigure}[b]{.24\linewidth}
\includegraphics[width=\linewidth]{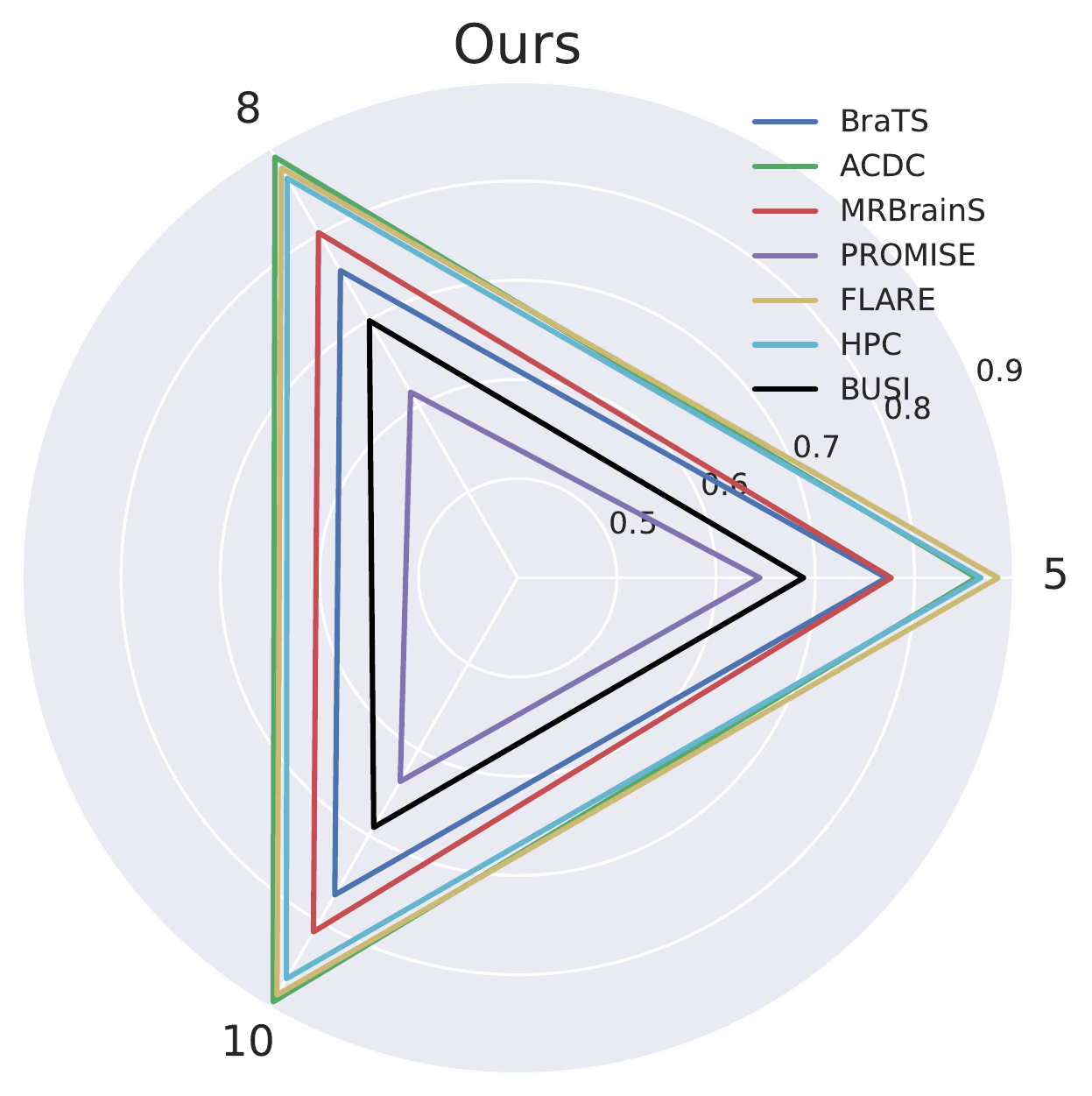}
\end{subfigure}
\begin{subfigure}[b]{.24\linewidth}
\includegraphics[width=\linewidth]{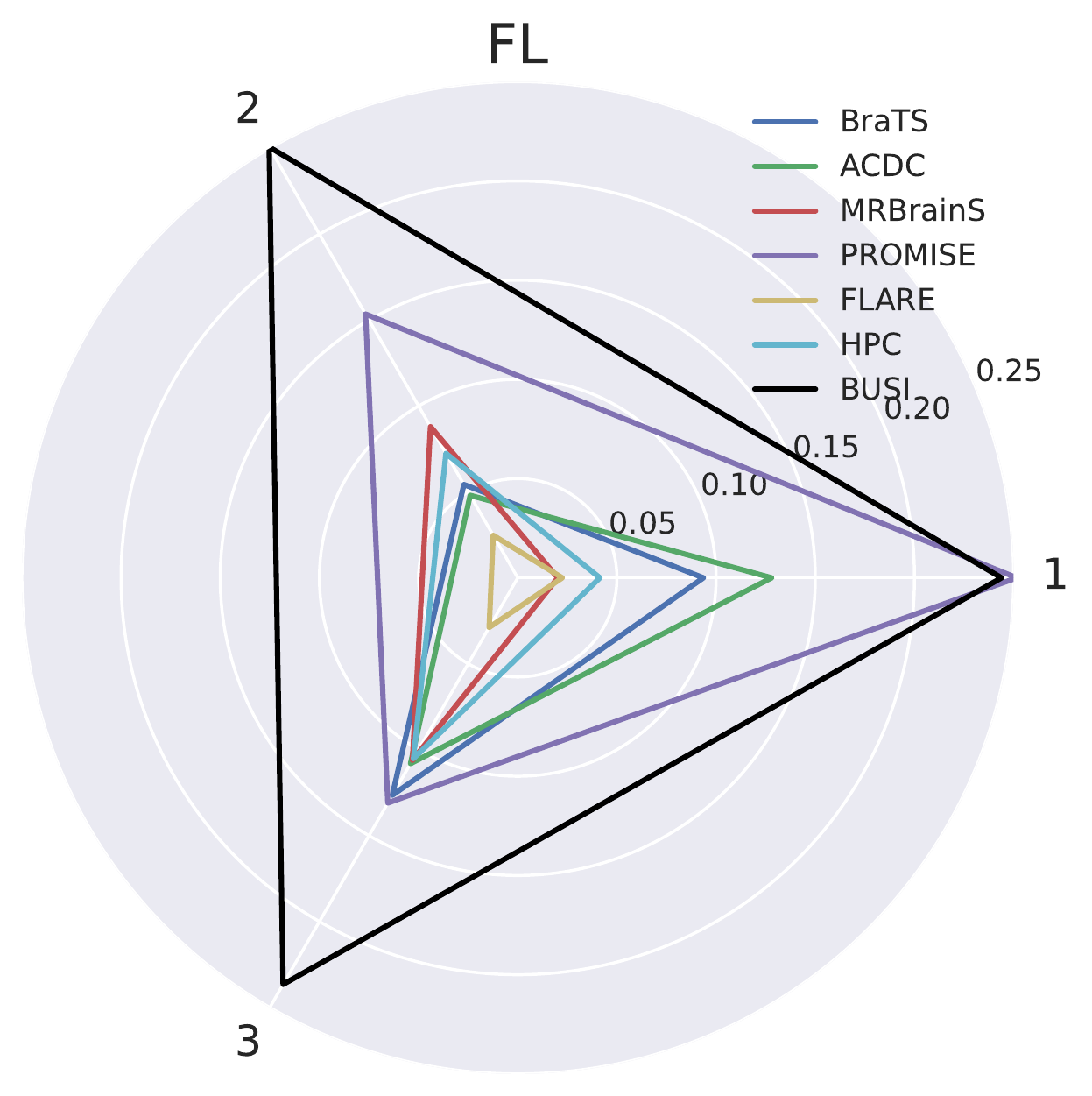}
\end{subfigure}
\begin{subfigure}[b]{.24\linewidth}
\includegraphics[width=\linewidth]{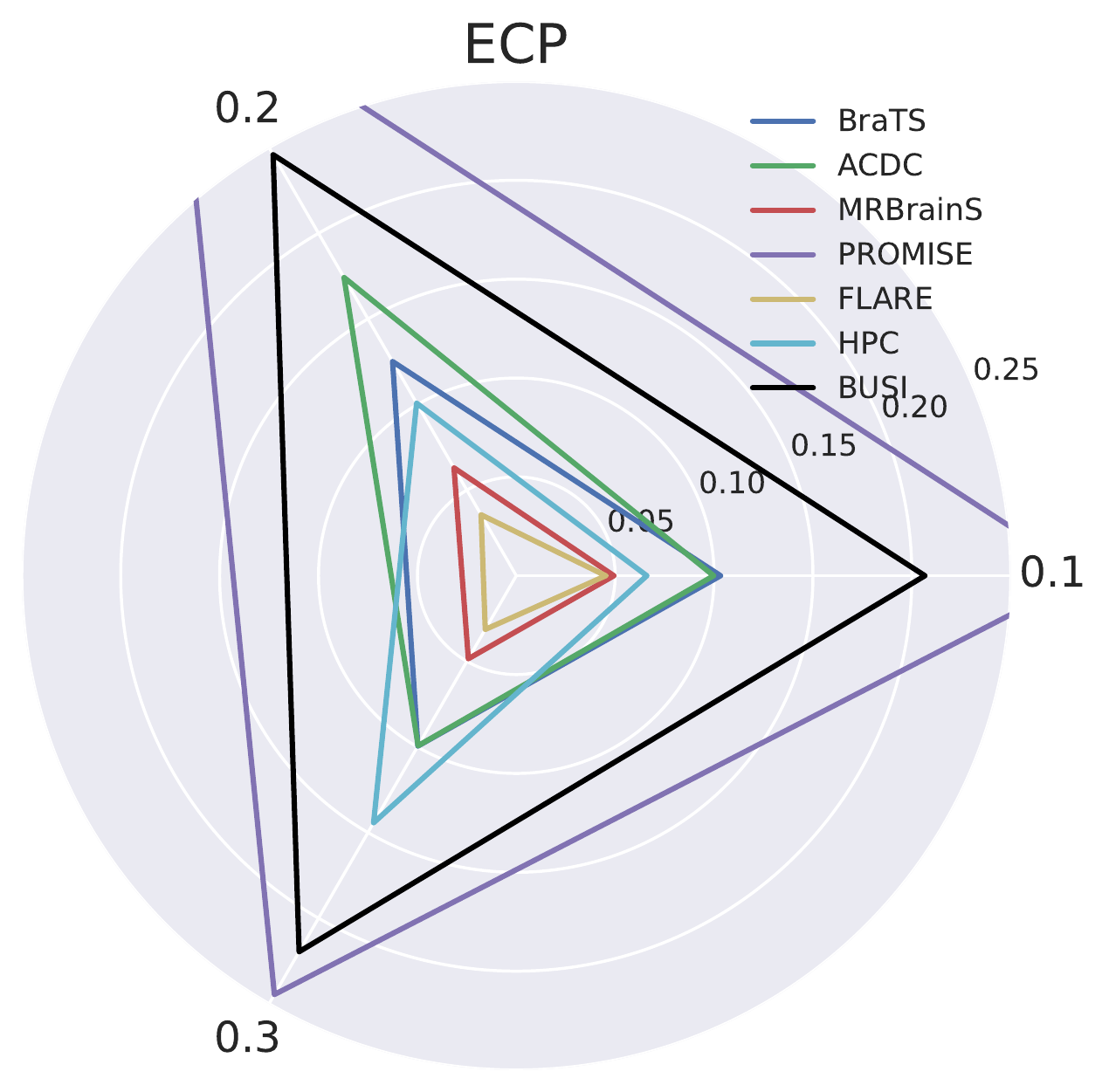}
\end{subfigure}
\begin{subfigure}[b]{.24\linewidth}
\includegraphics[width=\linewidth]{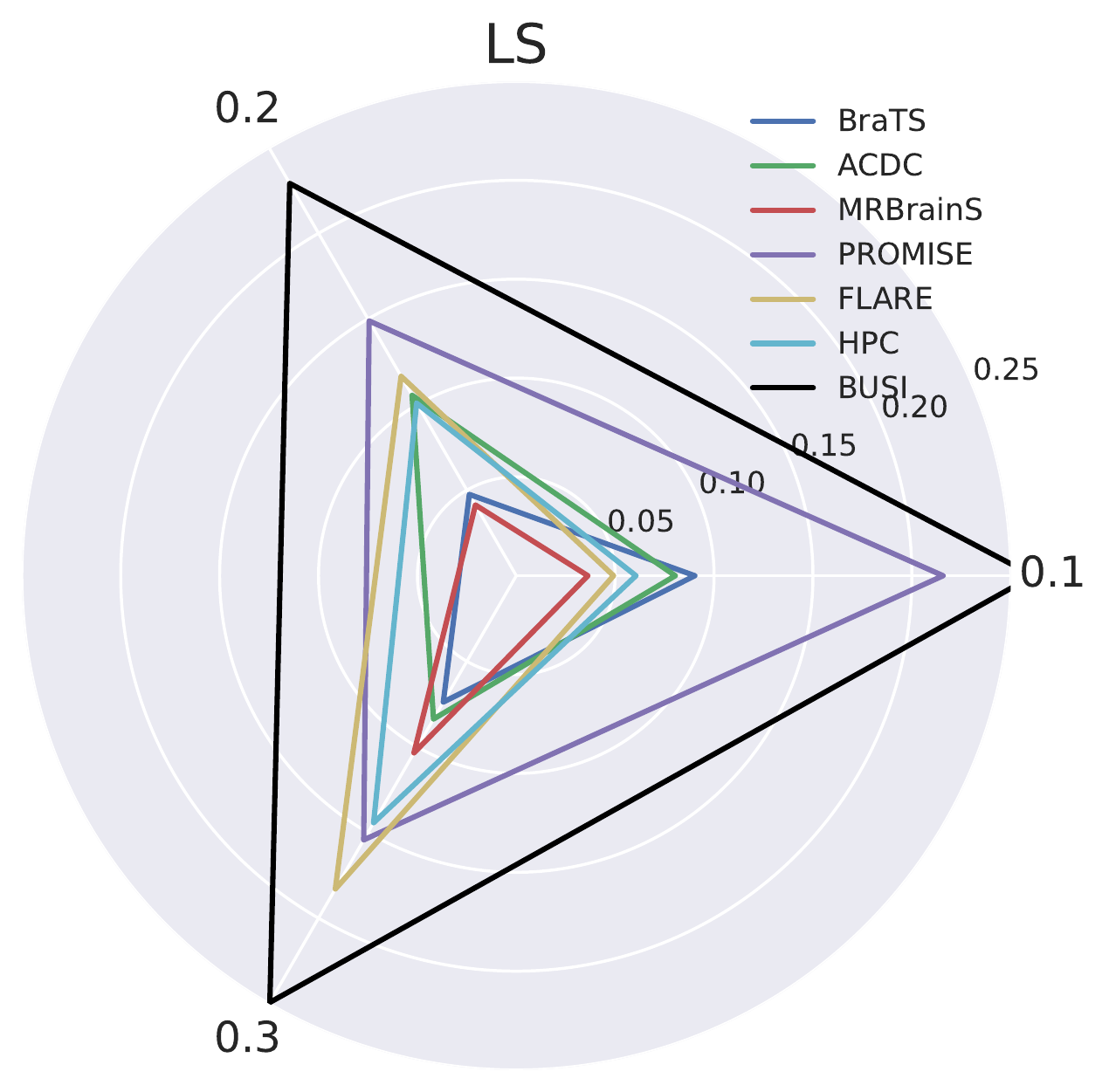}
\end{subfigure}
\begin{subfigure}[b]{.24\linewidth}
\includegraphics[width=\linewidth]{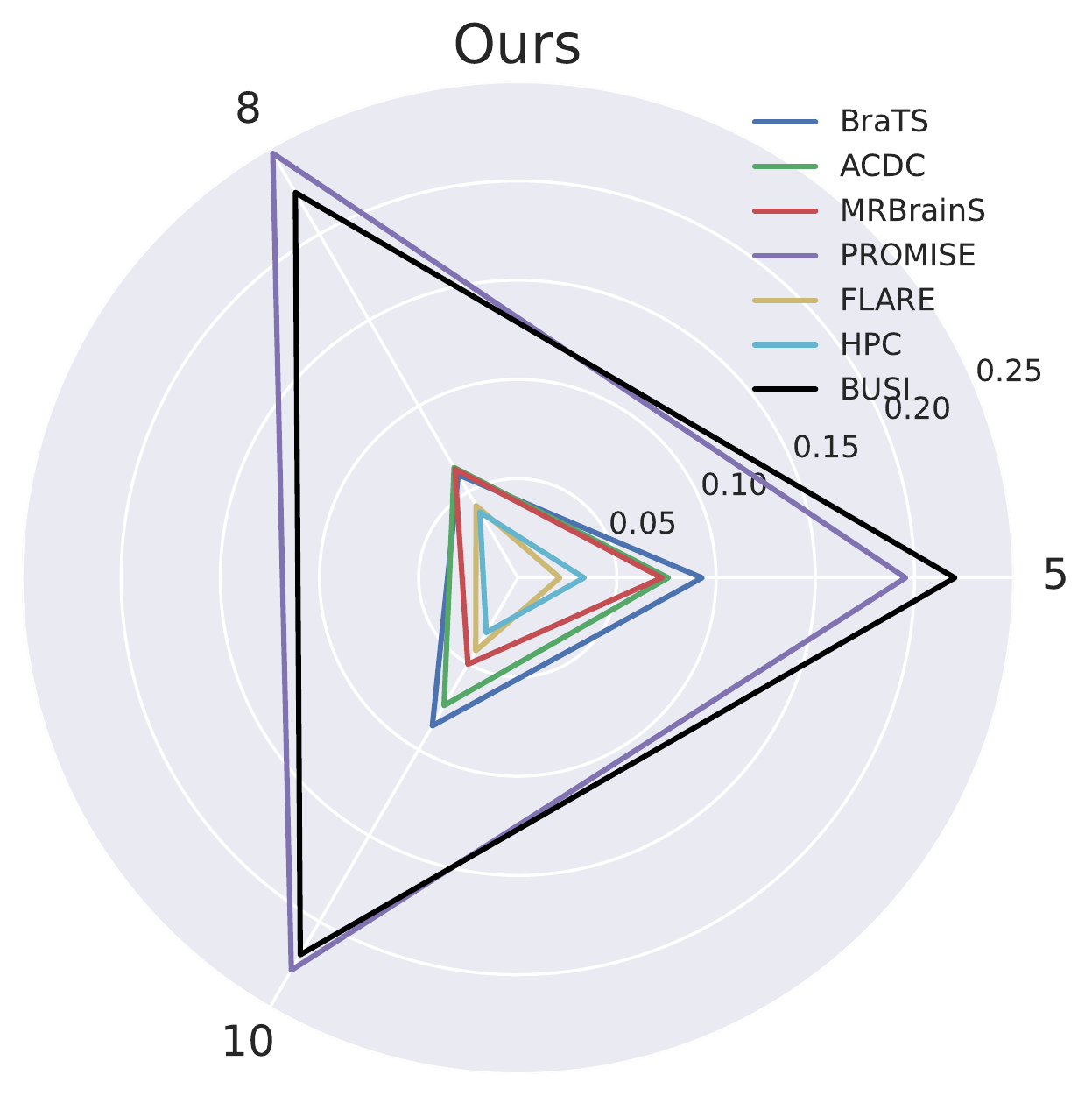}
\end{subfigure}
\caption{\rev{\textbf{Sensibility to hyperparameters across datasets.} For each method, we use the standard hyperparameters used in the literature and compare its variation across different datasets. The discriminative performance (DSC) is reported in the \textit{top row}, whereas the calibration analysis (ECE) is depicted in the \textit{bottom row}.}}
 \label{fig:rob_hyperparameter}
\end{figure*}

\subsubsection{\rev{Calibration and discriminative performance under distribution shift.}}

\rev{There have been recent empirical studies \citep{ovadia2019can,minderer2021revisiting} on the robustness of calibration models under distribution shift. In particular, \citep{minderer2021revisiting} explores out-of-distribution calibration by resorting to ImageNet-C \citep{hendrycks2018benchmarking}, a computer vision dataset that contains images that have been synthetically corrupted, for example by including Gaussian noise. Inspired by these works, we now assess the robustness of our model 
in the presence of domain drift. To do this, we added Gaussian noise to the images on the testing set, with sigma values ranging from 0 to 0.05 with an increment of 0.01. From the plots in Figure \ref{fig:rob_noise} we can clearly observe that models trained with our objective function are less sensitive to noise, compared to prior state-of-the-art methods. More concretely, on the PROMISE dataset, the discriminative and calibration performance of our approach remains almost invariant to image perturbations with different levels of Gaussian noise. Furthermore, while the results obtained by our method in the MRBrainS data are affected by noise, its performance degradation is significantly lower than nearly all previous approaches, being on par with the focal loss. Nonetheless, it is noteworthy to mention that despite the relative decrease in performance is similar between the proposed method and FL, their global performance differences are substantially large (e.g., ~6-8\% difference in DSC). Based on these results we can argue that the proposed method delivers higher performing models that are, in addition, more robust to distributional drifts produced by Gaussian noise. }

\subsubsection{\rev{On the impact of hyperparameters}}
\rev{
We now assess the sensitivity of each model to the choice of the hyperparameters on each dataset. 
We stress that, for each method, we have selected a range of common values used in the literature. In particular, $\gamma$ is set to 1.0, 2.0 and 3.0 in Focal loss, $\lambda$ is fixed to 0.1, 0.2 and 0.3 in ECP and Label smoothing, whereas the margin values in our method are set to 5.0, 8.0 and 10. 
The discriminative (DSC) and calibration (ECE) performances obtained across the different hyperparameter values are depicted in Fig. \ref{fig:rob_hyperparameter}. 
From this figure, it can be easily observed that, while prior approaches are very sensitive to the value of their balancing term, our method is significantly more robust to these changes. 
For example, the discriminative performance is drastically affected in both ECP and LS across several datasets when changing the value of the balancing term from 0.1 and 0.2 to 0.2 and 0.3, respectively. On the other hand, this phenomenon is more pronounced in the calibration metrics, where FL, ECP and LS show much higher variations than the proposed approach. A potential drawback that can be extracted from these findings is that, in order to get a well calibrated and high performance model, prior approaches might require multiple training iterations to find a satisfactory compromise. Furthermore, we believe that these large variations indicate that differences in the data --e.g., image contrast, target size and heterogeneity, or class distribution-- might have a different impact on these losses, entangling the convergence of models trained with these terms.}

\begin{figure}[h!]
     \centering
     \begin{subfigure}[b]{0.49\linewidth}
         \centering
         \includegraphics[width=\linewidth]{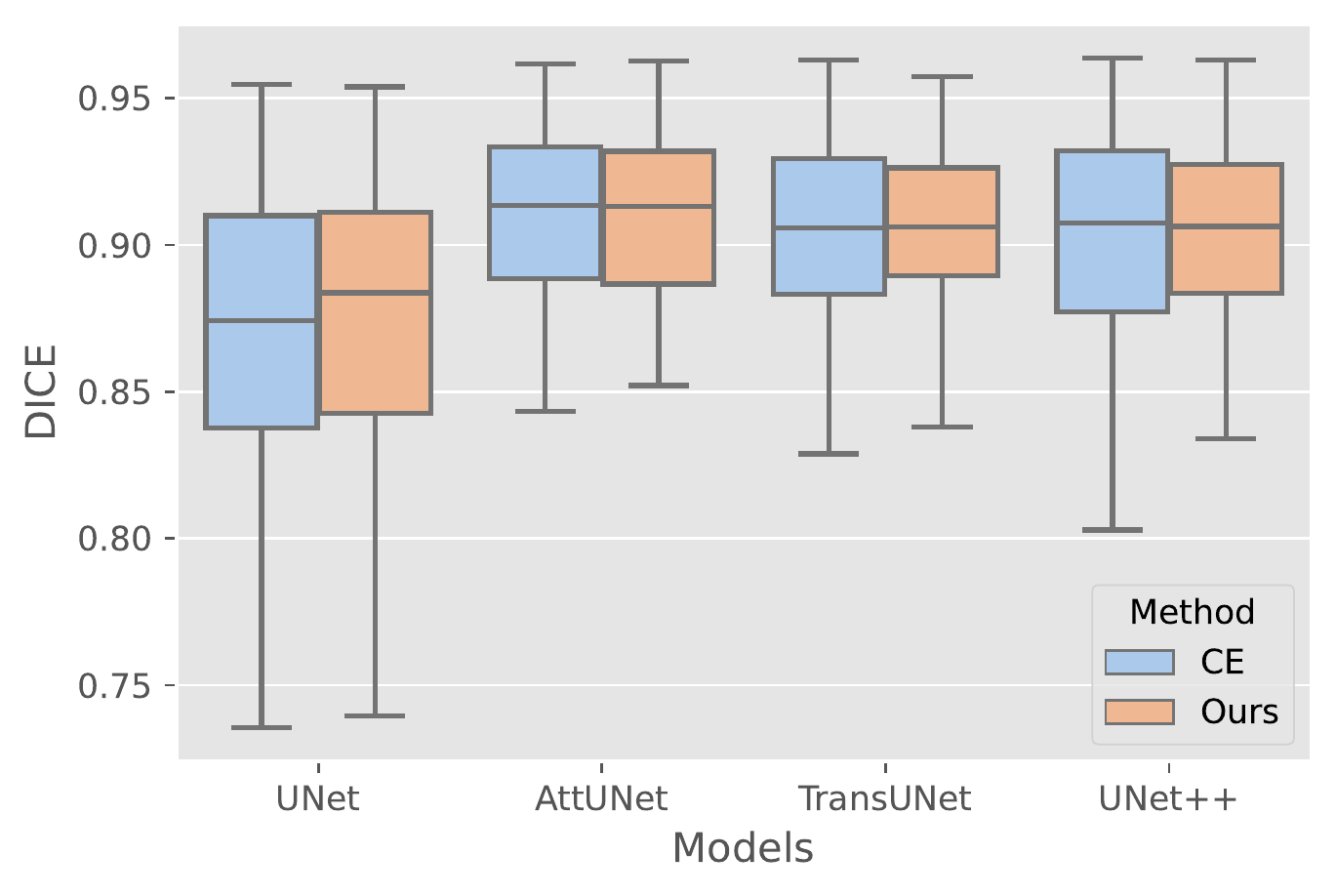}
     \end{subfigure}
     \begin{subfigure}[b]{0.49\linewidth}
         \centering
         \includegraphics[width=\linewidth]{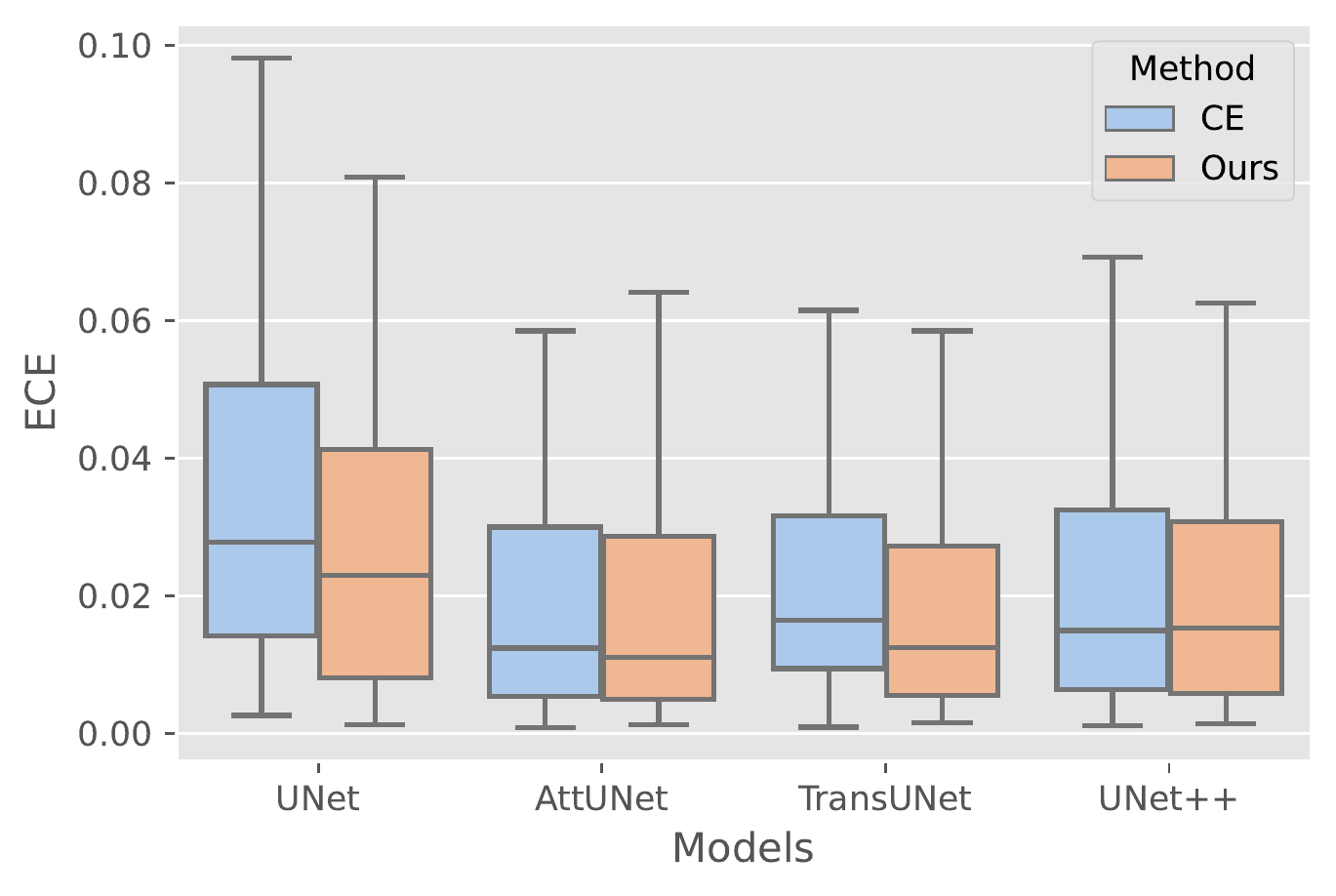}
     \end{subfigure}
        \caption{\revf{\textbf{Robustness to segmentation backbone}, which evaluates the standard cross-entropy and the proposed model on the FLARE segmentation benchmark.}}
        \label{fig:rob_backbone1}
\end{figure}

\subsubsection{\rev{Robustness to backbone}}
In this experiment, we evaluate the proposed loss on several standard medical image segmentation networks, including: AttUNet \citep{attunet}, TransUNet \citep{transunet}, and UNet++ \citep{unetpp}. For this study, we consider the FLARE dataset due to its larger number of classes. The quantitative comparison of CE and our method for these backbones is presented in Fig. \ref{fig:rob_backbone1}, from which it can be inferred that, irrespective of the backbone used, our method is capable of consistently achieving better model calibration compared to the popular cross-entropy loss, while yielding at par performance in the discrimination task. We can therefore say that the proposed term \revf{can be directly plugged into any segmentation network}, and the improvement observed is consistent across different models.

\begin{figure*}[h!]
    \begin{center}
    \includegraphics[width=.95\linewidth]{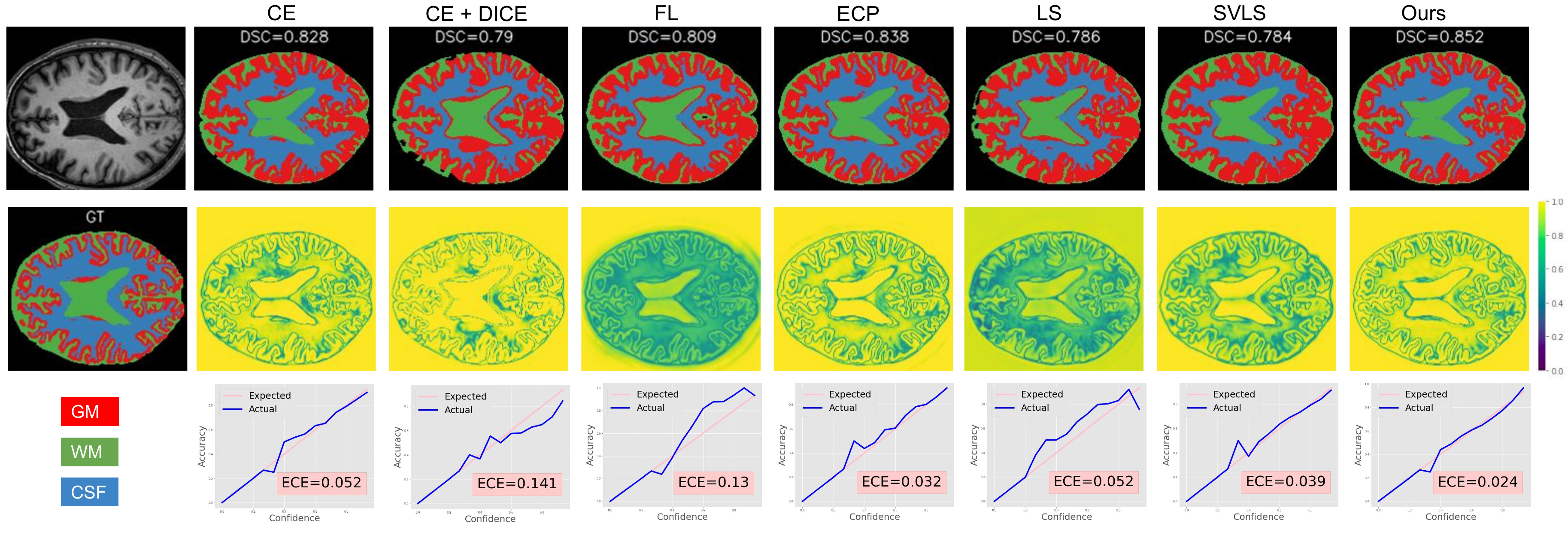}
    \end{center}
    \vspace{-1 em}
    \caption{\rev{Qualitative results on MRBrainS dataset for different methods. In particular, we show the original image and the corresponding segmentation masks provided by each method (\textit{top row}), the ground-truth (GT) mask followed by maximum confidence score of each method (\textit{middle row}) and the respective reliability plots (\textit{bottom row}). Methods from left to right: CE, CE+DICE, FL, ECP, LS, SVLS, Ours}. }
     \label{fig:qualitative}
\end{figure*}

In this experiment, we evaluate the proposed loss on several standard medical image segmentation networks, including: AttUNet \citep{attunet}, TransUNet \citep{transunet}, and UNet++ \citep{unetpp}. For this study, we consider the FLARE dataset due to its larger number of classes. The quantitative comparison of CE and our method for these backbones is presented in Fig. \ref{fig:rob_backbone1}, from which it can be inferred that, irrespective of the backbone used, our method is capable of consistently achieving better model calibration compared to the popular cross-entropy loss, while yielding at par performance in the discrimination task. We can therefore say that the proposed term \revf{can be directly plugged into any segmentation network}, and the improvement observed is consistent across different models.

\subsubsection{\rev{Qualitative results and reliability diagrams}}

\rev{Figure \ref{fig:qualitative} depicts the predicted segmentation masks (\textit{top}), confidence maps (\textit{middle}) and their corresponding reliability plots (\textit{bottom}) on one subject across the different methods. While the segmentation masks reveal several differences in terms of discriminative performance, the confidence maps present more interesting observations. Note that, as highlighted in prior works \citep{liu2021devil}, better calibrated models should show better edge sharpness, matching the expected property that the model should be less confident at the boundaries, whereas yielding more confident predictions in inner target regions. First, we can observe that adding the DSC loss term substantially degrades the confidence map compared to its single CE loss counterpart. In particular, 
the CE+DSC compounded loss tends to produce smoother edges, in terms of confidence, which is derived from worst calibrated models. Furthermore, while it can increase the confidence of predictions in several inner object regions, it decreases this score in others. In addition, we can clearly observe that the remaining analyzed methods provide a diverse span of confidence estimates, with several models providing highly unconfident inner regions (e.g., FL \citep{mukhoti2020calibrating} and LS \citep{szegedy2016rethinking}). In contrast, our method yields confidence estimates that are sharp in the edges and low in within-region pixels, as expected in a well-calibrated model. These visual findings are supported by the reliability diagrams. Indeed, our model yields the best reliability diagram, as the ECE curves are closer to the diagonal, indicating that the predicted probabilities serve as a good estimate of the correctness of the prediction. }

\subsubsection{\revf{Choice of the penalty.}}

\revf{In this work we have presented a unified constrained optimization perspective of existing calibration methods, showing that they can be seen as approximations of a linear penalty for imposing the constraint $\textbf{d(l)}=\textbf{0}$. In order to show that the improvement of the proposed formulation comes from the relaxation of this constraint, which has important limitations, we selected a linear penalty, similarly to the implicit underlying mechanism of LS, FL and ECP. Nevertheless, in this section we now address the question of whether we can further improve these results by employing other penalties to enforce the proposed constraint. In particular, we evaluate both the discriminative and calibration performance of our model when a quadratic penalty, i.e., $L_2$, is used to impose the constraint in Eq. \ref{eq:our-constraint}. Results in Table \ref{table:l1-vs-l2} show that, even though both penalties behave similarly in terms of segmentation, the model trained with a quadratic penalty is typically worse calibrated. We argue that these differences might be due to the more aggressive behaviour of quadratic penalties when the constraint is not satisfied, which may eventually lead to near-to-uninformative solutions, similar to those obtained by FL, LS and ECP. Furthermore, we would like to highlight that in this experiment the margin $m$ was fine-tuned for the $L_2$ penalty, whereas its controlling weight remained the same as in the $L_1$ term. Thus, further optimizing the penalty weight might alleviate its aggressive performance on large violations, potentially leading to superior performance of the proposed MbLS loss when the constraint is enforced via a quadratic penalty.}

\begin{table}[]
\centering
\footnotesize
\begin{tabular}{l|cccc}
\toprule
      & \multicolumn{2}{c}{\revf{L1}}        & \multicolumn{2}{c}{\revf{L2}} \\
        \midrule
      & \revf{DSC} & \revf{ECE $\downarrow$} &\revf{ DSC}        & \revf{ECE $\downarrow$}       \\
      \midrule
\revf{ACDC}  &  \revf{0.875}   &   \revf{0.061}   &  \revf{0.874}     &    \revf{ 0.064}      \\
\revf{FLARE} &  \revf{0.871} &    \revf{0.038}     & \revf{0.868}   &  \revf{0.031}        \\
\revf{BRATS} & \revf{0.854}    &     \revf{0.101}     &  \revf{0.845}  &     \revf{0.101}  \\
\revf{PROMISE}   &\revf{0.583}&     \revf{0.232}   & \revf{0.549} &  \revf{0.279}  \\
\revf{HPC}   & \revf{0.867}    &    \revf{0.033}   &   \revf{0.864}  & \revf{0.042} \\
\revf{BUSI}  &  \revf{0.685}   &  \revf{0.193}  & \revf{0.673} & \revf{0.197}  \\
\bottomrule
\end{tabular}
\caption{\revf{Quantitative comparison across datasets of different penalty terms to impose the constraint \textbf{d(l)} $\leq$ \textbf{m}}.}
\label{table:l1-vs-l2}
\end{table}

\subsubsection{\revf{Impact of MbLS on the CE + Dice loss}}

\begin{figure}[h!]
     \centering
     \begin{subfigure}[b]{0.49\linewidth}
         \centering
         \includegraphics[width=\linewidth]{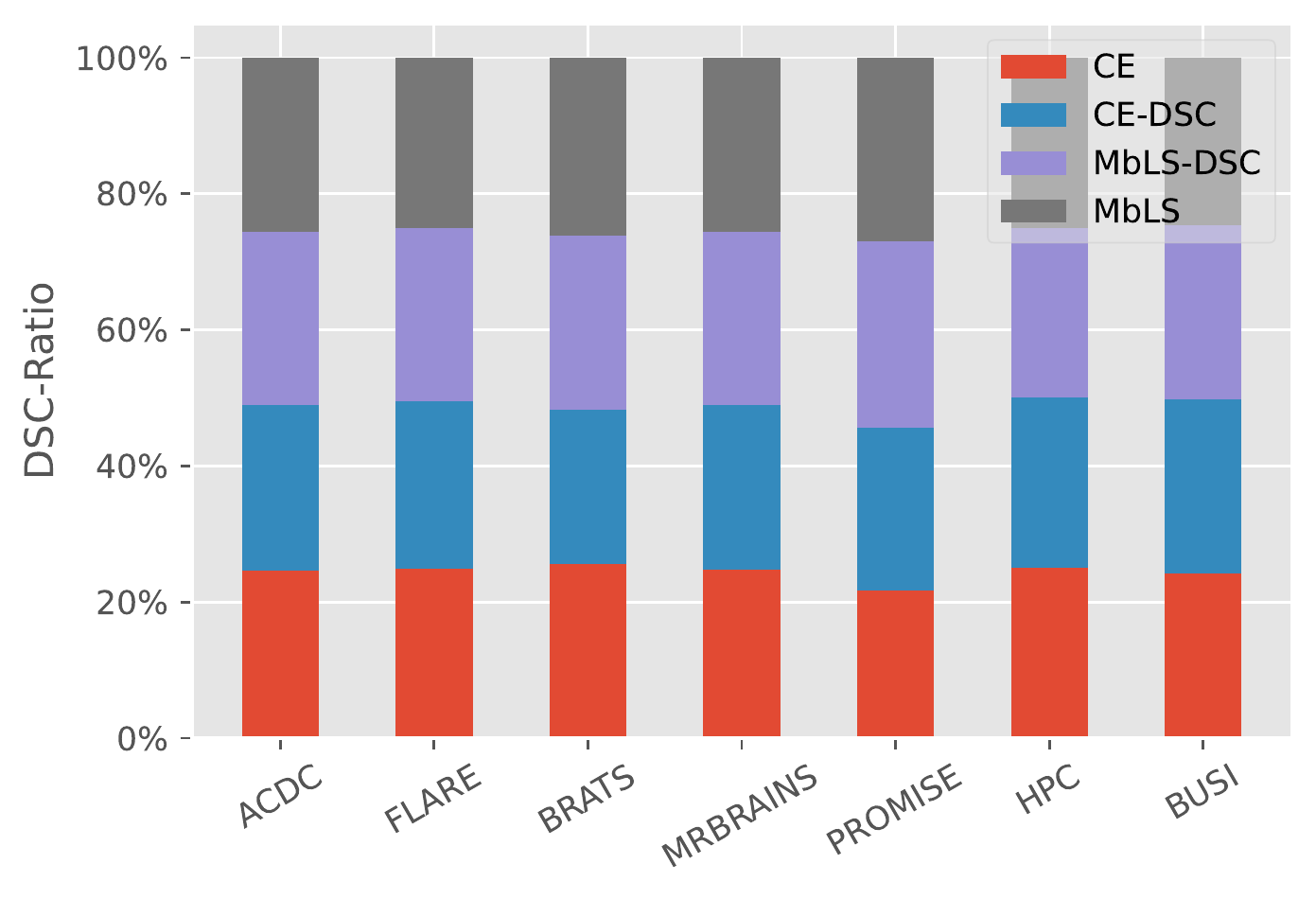}
     \end{subfigure}
     \begin{subfigure}[b]{0.49\linewidth}
         \centering
         \includegraphics[width=\linewidth]{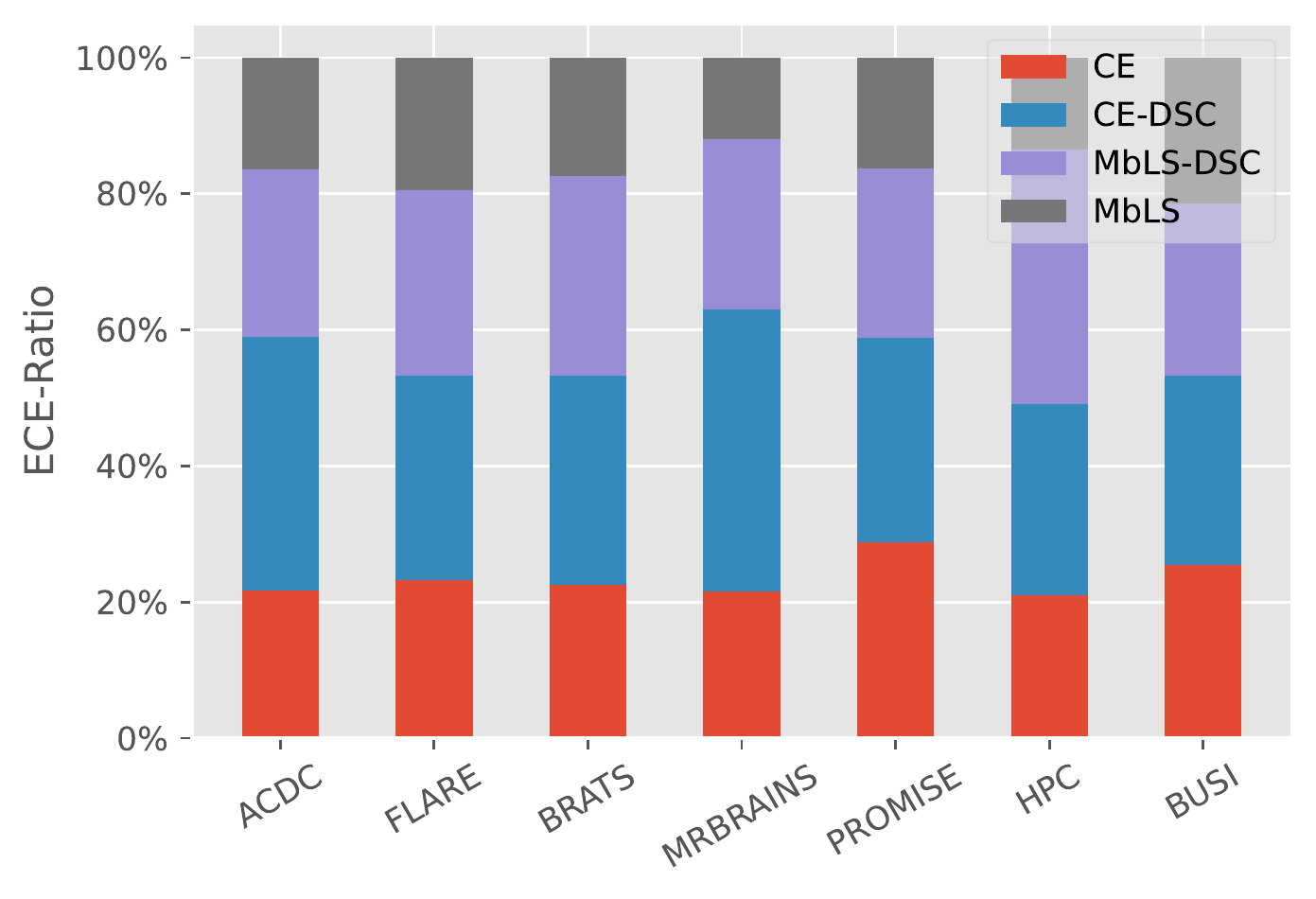}
     \end{subfigure}
    \caption{\revf{\textbf{Impact of MbLS on the DSC loss.} Normalized stacked bar plots to assess the impact of the proposed MbLS on the popular CE+DSC segmentation loss. Discriminative performance in terms of DSC is depicted in the \textit{top} (the higher the ratio the better), whereas calibration is assessed in terms of ECE in the \textit{bottom} (the lower the ratio the better).}}
    \label{fig:ce_dice_mbls}
\end{figure}

\revf{The main goal of this work is to present existing calibration losses from a constrained optimization perspective, highlight their weaknesses and propose a potential solution to overcome the identified limitations. Furthermore, we stress that existing calibration losses do not include compounded losses that integrate segmentation terms, such as the Dice loss. Nevertheless, given the popularity of this joint learning objective in medical image segmentation, we assess in this section the impact of integrating the proposed constrained term into the duple CE + DSC. In particular, to better understand the impact of the proposed MbLS loss, as well as DSC loss, we depict the results for the standard CE, CE + DSC, MbLS and MbLS + DSC in Fig \ref{fig:ce_dice_mbls}. The stacked normalized plots show that while these methods result in similar discriminative performance, the differences in calibration are more noticeable. More concretely, and as we demonstrated empirically in Table \ref{tab:main-cal}, models trained with the joint CE + DSC loss see their calibration performance degrade compared to the standard CE objective. By coupling the proposed approach with the DSC loss ($MbLS + DSC$) this degradation can be reduced thanks to the proposed penalty term. Nevertheless, the ECE results achieved by this joint term are significantly higher than those obtained by the proposed approach, which does not include the DSC loss. These findings demonstrate that \textit{i}) the proposed MbLS can improve the calibration performance of the popular CE + DSC segmentation loss, and \textit{ii)} despite the improvement in discrimination performance, losses integrating the DSC loss term present significant deficiencies to deliver well-calibrated models.}

\section{\rev{Conclusion}}
\rev{Despite the popularity of network calibration in a broad span of applications, the connection between state-of-the-art calibration losses remains unexplored, and their impact on segmentation networks, particularly in the medical field, has largely been overlooked. In this work, we have demonstrated that these popular losses are closely related from a constrained optimization perspective, whose implicit or explicit constraints lead to non-informative solutions, preventing the model predictions to reach the best compromise between discriminative and calibration performance. To overcome this issue, we proposed a simple solution that integrates an inequality constraint into the main learning objective, which imposes a controlled margin on the logit distances. 
Through an extensive empirical evaluation, which contains multiple popular segmentation benchmarks, we have assessed the discriminative and calibration performance of state-of-the-art calibration losses in the important task of medical image segmentation. The results highlight several important benefits of the proposed loss. First, it achieves consistent improvements over state-of-the-art calibration and segmentation losses, both in terms of discriminative and calibration performance. Second, the proposed model is much less sensitive to hyperparameters changes compared to prior losses, which reduces the training time to find a satisfactory compromise between discrimination and calibration tasks. In addition, the empirical observations support our hypothesis that the suboptimal supervision delivered by the standard cross-entropy loss likely results in poorly calibrated models, as model trained with this loss tend to produce largest logit differences. Thus, we advocate that the proposed loss term should be preferred to train models that provide higher discriminative performance, while yet delivering accurate uncertainty estimates.}

\section*{Acknowledgements}
This work is supported by the National Science and Engineering Research Council of Canada (NSERC), via its Discovery Grant program and FRQNT through the Research Support for New Academics program. We also thank Calcul Quebec and Compute Canada. Adrian Galdran was funded by a Marie Sklodowska-Curie Fellowship (No 892297).


\bibliographystyle{model2-names.bst}\biboptions{authoryear}
\bibliography{main}

\end{document}

%% file: notation.tex
\usepackage{amsmath,amsfonts,bm}

\def\eqref#1{equation~\ref{#1}}

\def\1{\bm{1}}

\DeclareMathAlphabet{\mathsfit}{\encodingdefault}{\sfdefault}{m}{sl}
\SetMathAlphabet{\mathsfit}{bold}{\encodingdefault}{\sfdefault}{bx}{n}

\DeclareMathOperator*{\argmax}{arg\,max}

\newcommand{\ceq}{\stackrel{\mathclap{\normalfont\mbox{c}}}{=}}
\newcommand{\cleq}{\stackrel{\mathclap{\normalfont\mbox{c}}}{\leq}}

\newcommand{\dd}{\mathbf{d}}
\newcommand{\xx}{\mathbf{x}}
\newcommand{\llll}{\mathbf{l}}

\newcommand{\yy}{\mathbf{y}}

\newcommand{\s}{\mathbf{s}}
\newcommand{\sss}{\mathbf{s}}
\newcommand{\0}{\mathbf{0}}

\newtheorem{proposition}{Proposition}

%% file: main.bbl
\begin{thebibliography}{51}
\expandafter\ifx\csname natexlab\endcsname\relax\def\natexlab#1{#1}\fi
\providecommand{\url}[1]{\texttt{#1}}
\providecommand{\href}[2]{#2}
\providecommand{\path}[1]{#1}
\providecommand{\DOIprefix}{doi:}
\providecommand{\ArXivprefix}{arXiv:}
\providecommand{\URLprefix}{URL: }
\providecommand{\Pubmedprefix}{pmid:}
\providecommand{\doi}[1]{\href{http://dx.doi.org/#1}{\path{#1}}}
\providecommand{\Pubmed}[1]{\href{pmid:#1}{\path{#1}}}
\providecommand{\bibinfo}[2]{#2}
\ifx\xfnm\relax \def\xfnm[#1]{\unskip,\space#1}\fi
\bibitem[{Al-Dhabyani et~al.(2020)Al-Dhabyani, Gomaa, Khaled and Fahmy}]{al2020dataset}
\bibinfo{author}{Al-Dhabyani, W.}, \bibinfo{author}{Gomaa, M.}, \bibinfo{author}{Khaled, H.}, \bibinfo{author}{Fahmy, A.}, \bibinfo{year}{2020}.
\newblock \bibinfo{title}{Dataset of breast ultrasound images}.
\newblock \bibinfo{journal}{Data in brief} \bibinfo{volume}{28}, \bibinfo{pages}{104863}.
\bibitem[{Antonelli et~al.(2022)Antonelli, Reinke, Bakas, Farahani, Kopp-Schneider, Landman, Litjens, Menze, Ronneberger, Summers et~al.}]{antonelli2022medical}
\bibinfo{author}{Antonelli, M.}, \bibinfo{author}{Reinke, A.}, \bibinfo{author}{Bakas, S.}, \bibinfo{author}{Farahani, K.}, \bibinfo{author}{Kopp-Schneider, A.}, \bibinfo{author}{Landman, B.A.}, \bibinfo{author}{Litjens, G.}, \bibinfo{author}{Menze, B.}, \bibinfo{author}{Ronneberger, O.}, \bibinfo{author}{Summers, R.M.}, et~al., \bibinfo{year}{2022}.
\newblock \bibinfo{title}{The medical segmentation decathlon}.
\newblock \bibinfo{journal}{Nature communications} \bibinfo{volume}{13}, \bibinfo{pages}{4128}.
\bibitem[{Bakas et~al.(2017)Bakas, Akbari, Sotiras, Bilello, Rozycki, Kirby, Freymann, Farahani and Davatzikos}]{Bakas2017AdvancingFeaturesJ}
\bibinfo{author}{Bakas, S.}, \bibinfo{author}{Akbari, H.}, \bibinfo{author}{Sotiras, A.}, \bibinfo{author}{Bilello, M.}, \bibinfo{author}{Rozycki, M.}, \bibinfo{author}{Kirby, J.S.}, \bibinfo{author}{Freymann, J.B.}, \bibinfo{author}{Farahani, K.}, \bibinfo{author}{Davatzikos, C.}, \bibinfo{year}{2017}.
\newblock \bibinfo{title}{Advancing the cancer genome atlas glioma mri collections with expert segmentation labels and radiomic features}.
\newblock \bibinfo{journal}{Scientific data} \bibinfo{volume}{4}, \bibinfo{pages}{1--13}.
\bibitem[{Bakas et~al.(2018)Bakas, Reyes, Jakab, Bauer, Rempfler, Crimi, Shinohara, Berger, Ha, Rozycki et~al.}]{Bakas2018IdentifyingChallengeJ}
\bibinfo{author}{Bakas, S.}, \bibinfo{author}{Reyes, M.}, \bibinfo{author}{Jakab, A.}, \bibinfo{author}{Bauer, S.}, \bibinfo{author}{Rempfler, M.}, \bibinfo{author}{Crimi, A.}, \bibinfo{author}{Shinohara, R.T.}, \bibinfo{author}{Berger, C.}, \bibinfo{author}{Ha, S.M.}, \bibinfo{author}{Rozycki, M.}, et~al., \bibinfo{year}{2018}.
\newblock \bibinfo{title}{Identifying the best machine learning algorithms for brain tumor segmentation, progression assessment, and overall survival prediction in the brats challenge}.
\newblock \bibinfo{journal}{arXiv preprint arXiv:1811.02629} .
\bibitem[{Bernard et~al.(2018)Bernard, Lalande, Zotti, Cervenansky, Yang, Heng, Cetin, Lekadir, Camara, Ballester et~al.}]{bernard2018deep}
\bibinfo{author}{Bernard, O.}, \bibinfo{author}{Lalande, A.}, \bibinfo{author}{Zotti, C.}, \bibinfo{author}{Cervenansky, F.}, \bibinfo{author}{Yang, X.}, \bibinfo{author}{Heng, P.A.}, \bibinfo{author}{Cetin, I.}, \bibinfo{author}{Lekadir, K.}, \bibinfo{author}{Camara, O.}, \bibinfo{author}{Ballester, M.A.G.}, et~al., \bibinfo{year}{2018}.
\newblock \bibinfo{title}{Deep learning techniques for automatic {MRI} cardiac multi-structures segmentation and diagnosis: is the problem solved?}
\newblock \bibinfo{journal}{IEEE TMI} \bibinfo{volume}{37}, \bibinfo{pages}{2514--2525}.
\bibitem[{Bertsekas(1995)}]{Bertsekas95}
\bibinfo{author}{Bertsekas, D.}, \bibinfo{year}{1995}.
\newblock \bibinfo{title}{Nonlinear Programming}.
\newblock \bibinfo{publisher}{Athena Scientific, Belmont, MA}.
\bibitem[{Blundell et~al.(2015)Blundell, Cornebise, Kavukcuoglu and Wierstra}]{blundell2015weight}
\bibinfo{author}{Blundell, C.}, \bibinfo{author}{Cornebise, J.}, \bibinfo{author}{Kavukcuoglu, K.}, \bibinfo{author}{Wierstra, D.}, \bibinfo{year}{2015}.
\newblock \bibinfo{title}{Weight uncertainty in neural network}, in: \bibinfo{booktitle}{ICML}.
\bibitem[{Chen et~al.(2021)Chen, Lu, Yu, Luo, Adeli, Wang, Lu, Yuille and Zhou}]{transunet}
\bibinfo{author}{Chen, J.}, \bibinfo{author}{Lu, Y.}, \bibinfo{author}{Yu, Q.}, \bibinfo{author}{Luo, X.}, \bibinfo{author}{Adeli, E.}, \bibinfo{author}{Wang, Y.}, \bibinfo{author}{Lu, L.}, \bibinfo{author}{Yuille, A.L.}, \bibinfo{author}{Zhou, Y.}, \bibinfo{year}{2021}.
\newblock \bibinfo{title}{Transunet: Transformers make strong encoders for medical image segmentation}.
\newblock \bibinfo{journal}{arXiv preprint arXiv:2102.04306} .
\bibitem[{Ding et~al.(2021)Ding, Han, Liu and Niethammer}]{ding2021local}
\bibinfo{author}{Ding, Z.}, \bibinfo{author}{Han, X.}, \bibinfo{author}{Liu, P.}, \bibinfo{author}{Niethammer, M.}, \bibinfo{year}{2021}.
\newblock \bibinfo{title}{Local temperature scaling for probability calibration}, in: \bibinfo{booktitle}{Proceedings of the IEEE/CVF International Conference on Computer Vision}, pp. \bibinfo{pages}{6889--6899}.
\bibitem[{Fort et~al.(2019)Fort, Hu and Lakshminarayanan}]{fort2019deep}
\bibinfo{author}{Fort, S.}, \bibinfo{author}{Hu, H.}, \bibinfo{author}{Lakshminarayanan, B.}, \bibinfo{year}{2019}.
\newblock \bibinfo{title}{Deep ensembles: A loss landscape perspective}.
\newblock \bibinfo{journal}{arXiv preprint arXiv:1912.02757} .
\bibitem[{Gal and Ghahramani(2016)}]{gal2016dropout}
\bibinfo{author}{Gal, Y.}, \bibinfo{author}{Ghahramani, Z.}, \bibinfo{year}{2016}.
\newblock \bibinfo{title}{Dropout as a bayesian approximation: Representing model uncertainty in deep learning}, in: \bibinfo{booktitle}{ICML}.
\bibitem[{Guo et~al.(2017)Guo, Pleiss, Sun and Weinberger}]{guo2017calibration}
\bibinfo{author}{Guo, C.}, \bibinfo{author}{Pleiss, G.}, \bibinfo{author}{Sun, Y.}, \bibinfo{author}{Weinberger, K.Q.}, \bibinfo{year}{2017}.
\newblock \bibinfo{title}{On calibration of modern neural networks}, in: \bibinfo{booktitle}{ICML}.
\bibitem[{Hendrycks and Dietterich(2018)}]{hendrycks2018benchmarking}
\bibinfo{author}{Hendrycks, D.}, \bibinfo{author}{Dietterich, T.}, \bibinfo{year}{2018}.
\newblock \bibinfo{title}{Benchmarking neural network robustness to common corruptions and perturbations}, in: \bibinfo{booktitle}{International Conference on Learning Representations}.
\bibitem[{Hern{\'a}ndez-Lobato and Adams(2015)}]{hernandez2015probabilistic}
\bibinfo{author}{Hern{\'a}ndez-Lobato, J.M.}, \bibinfo{author}{Adams, R.}, \bibinfo{year}{2015}.
\newblock \bibinfo{title}{Probabilistic backpropagation for scalable learning of bayesian neural networks}, in: \bibinfo{booktitle}{ICML}.
\bibitem[{Islam and Glocker(2021)}]{islam2021spatially}
\bibinfo{author}{Islam, M.}, \bibinfo{author}{Glocker, B.}, \bibinfo{year}{2021}.
\newblock \bibinfo{title}{Spatially varying label smoothing: Capturing uncertainty from expert annotations}, in: \bibinfo{booktitle}{International Conference on Information Processing in Medical Imaging}, pp. \bibinfo{pages}{677--688}.
\bibitem[{Jena and Awate(2019)}]{jena2019bayesian}
\bibinfo{author}{Jena, R.}, \bibinfo{author}{Awate, S.P.}, \bibinfo{year}{2019}.
\newblock \bibinfo{title}{A bayesian neural net to segment images with uncertainty estimates and good calibration}, in: \bibinfo{booktitle}{International Conference on Information Processing in Medical Imaging}, pp. \bibinfo{pages}{3--15}.
\bibitem[{Jungo et~al.(2020)Jungo, Balsiger and Reyes}]{jungo2020analyzing}
\bibinfo{author}{Jungo, A.}, \bibinfo{author}{Balsiger, F.}, \bibinfo{author}{Reyes, M.}, \bibinfo{year}{2020}.
\newblock \bibinfo{title}{Analyzing the quality and challenges of uncertainty estimations for brain tumor segmentation}.
\newblock \bibinfo{journal}{Frontiers in neuroscience} \bibinfo{volume}{14}, \bibinfo{pages}{282}.
\bibitem[{Karimi and Gholipour(2022)}]{karimi2022improving}
\bibinfo{author}{Karimi, D.}, \bibinfo{author}{Gholipour, A.}, \bibinfo{year}{2022}.
\newblock \bibinfo{title}{Improving calibration and out-of-distribution detection in deep models for medical image segmentation}.
\newblock \bibinfo{journal}{IEEE Transactions on Artificial Intelligence} .
\bibitem[{Kock et~al.(2021)Kock, Thielke, Chlebus and Meine}]{kock2021confidence}
\bibinfo{author}{Kock, F.}, \bibinfo{author}{Thielke, F.}, \bibinfo{author}{Chlebus, G.}, \bibinfo{author}{Meine, H.}, \bibinfo{year}{2021}.
\newblock \bibinfo{title}{Confidence histograms for model reliability analysis and temperature calibration}, in: \bibinfo{booktitle}{Medical Imaging with Deep Learning}.
\bibitem[{Lakshminarayanan et~al.(2017)Lakshminarayanan, Pritzel and Blundell}]{lakshminarayanan2016simple}
\bibinfo{author}{Lakshminarayanan, B.}, \bibinfo{author}{Pritzel, A.}, \bibinfo{author}{Blundell, C.}, \bibinfo{year}{2017}.
\newblock \bibinfo{title}{Simple and scalable predictive uncertainty estimation using deep ensembles}, in: \bibinfo{booktitle}{NeurIPS}.
\bibitem[{Larrazabal et~al.(2021)Larrazabal, Mart{\'\i}nez, Dolz and Ferrante}]{larrazabal2021orthogonal}
\bibinfo{author}{Larrazabal, A.J.}, \bibinfo{author}{Mart{\'\i}nez, C.}, \bibinfo{author}{Dolz, J.}, \bibinfo{author}{Ferrante, E.}, \bibinfo{year}{2021}.
\newblock \bibinfo{title}{Orthogonal ensemble networks for biomedical image segmentation}, in: \bibinfo{booktitle}{MICCAI}.
\bibitem[{Lin et~al.(2017)Lin, Goyal, Girshick, He and Doll{\'a}r}]{lin2017focal}
\bibinfo{author}{Lin, T.Y.}, \bibinfo{author}{Goyal, P.}, \bibinfo{author}{Girshick, R.}, \bibinfo{author}{He, K.}, \bibinfo{author}{Doll{\'a}r, P.}, \bibinfo{year}{2017}.
\newblock \bibinfo{title}{Focal loss for dense object detection}, in: \bibinfo{booktitle}{CVPR}.
\bibitem[{Litjens et~al.(2014)Litjens, Toth, van~de Ven, Hoeks, Kerkstra, van Ginneken, Vincent, Guillard, Birbeck, Zhang et~al.}]{litjens2014evaluation}
\bibinfo{author}{Litjens, G.}, \bibinfo{author}{Toth, R.}, \bibinfo{author}{van~de Ven, W.}, \bibinfo{author}{Hoeks, C.}, \bibinfo{author}{Kerkstra, S.}, \bibinfo{author}{van Ginneken, B.}, \bibinfo{author}{Vincent, G.}, \bibinfo{author}{Guillard, G.}, \bibinfo{author}{Birbeck, N.}, \bibinfo{author}{Zhang, J.}, et~al., \bibinfo{year}{2014}.
\newblock \bibinfo{title}{Evaluation of prostate segmentation algorithms for mri: the promise12 challenge}.
\newblock \bibinfo{journal}{Medical image analysis} \bibinfo{volume}{18}, \bibinfo{pages}{359--373}.
\bibitem[{Liu et~al.(2022)Liu, Ben~Ayed, Galdran and Dolz}]{liu2021devil}
\bibinfo{author}{Liu, B.}, \bibinfo{author}{Ben~Ayed, I.}, \bibinfo{author}{Galdran, A.}, \bibinfo{author}{Dolz, J.}, \bibinfo{year}{2022}.
\newblock \bibinfo{title}{The devil is in the margin: Margin-based label smoothing for network calibration}, in: \bibinfo{booktitle}{CVPR}.
\bibitem[{Louizos and Welling(2016)}]{louizos2016structured}
\bibinfo{author}{Louizos, C.}, \bibinfo{author}{Welling, M.}, \bibinfo{year}{2016}.
\newblock \bibinfo{title}{Structured and efficient variational deep learning with matrix gaussian posteriors}, in: \bibinfo{booktitle}{ICML}.
\bibitem[{Lukasik et~al.(2020)Lukasik, Bhojanapalli, Menon and Kumar}]{lukasik2020does}
\bibinfo{author}{Lukasik, M.}, \bibinfo{author}{Bhojanapalli, S.}, \bibinfo{author}{Menon, A.}, \bibinfo{author}{Kumar, S.}, \bibinfo{year}{2020}.
\newblock \bibinfo{title}{Does label smoothing mitigate label noise?}, in: \bibinfo{booktitle}{ICML}.
\bibitem[{Ma et~al.(2021)Ma, Zhang, Gu, Zhu, Ge, Zhang, An, Wang, Wang, Liu, Cao, Zhang, Liu, Wang, Li, He and Yang}]{AbdomenCT-1K}
\bibinfo{author}{Ma, J.}, \bibinfo{author}{Zhang, Y.}, \bibinfo{author}{Gu, S.}, \bibinfo{author}{Zhu, C.}, \bibinfo{author}{Ge, C.}, \bibinfo{author}{Zhang, Y.}, \bibinfo{author}{An, X.}, \bibinfo{author}{Wang, C.}, \bibinfo{author}{Wang, Q.}, \bibinfo{author}{Liu, X.}, \bibinfo{author}{Cao, S.}, \bibinfo{author}{Zhang, Q.}, \bibinfo{author}{Liu, S.}, \bibinfo{author}{Wang, Y.}, \bibinfo{author}{Li, Y.}, \bibinfo{author}{He, J.}, \bibinfo{author}{Yang, X.}, \bibinfo{year}{2021}.
\newblock \bibinfo{title}{Abdomenct-1k: Is abdominal organ segmentation a solved problem?}
\newblock \bibinfo{journal}{IEEE Transactions on Pattern Analysis and Machine Intelligence} \DOIprefix\doi{10.1109/TPAMI.2021.3100536}.
\bibitem[{Ma and Blaschko(2021)}]{Ma2021postrank}
\bibinfo{author}{Ma, X.}, \bibinfo{author}{Blaschko, M.B.}, \bibinfo{year}{2021}.
\newblock \bibinfo{title}{Meta-cal: Well-controlled post-hoc calibration by ranking}, in: \bibinfo{booktitle}{ICML}.
\bibitem[{Maier et~al.(2017)Maier, Menze, {von der Gablentz}, Häni, Heinrich, Liebrand, Winzeck, Basit, Bentley, Chen, Christiaens, Dutil, Egger, Feng, Glocker, Götz, Haeck, Halme, Havaei, Iftekharuddin, Jodoin, Kamnitsas, Kellner, Korvenoja, Larochelle, Ledig, Lee, Maes, Mahmood, Maier-Hein, McKinley, Muschelli, Pal, Pei, Rangarajan, Reza, Robben, Rueckert, Salli, Suetens, Wang, Wilms, Kirschke, Krämer, Münte, Schramm, Wiest, Handels and Reyes}]{MAIER2017250}
\bibinfo{author}{Maier, O.}, \bibinfo{author}{Menze, B.H.}, \bibinfo{author}{{von der Gablentz}, J.}, \bibinfo{author}{Häni, L.}, \bibinfo{author}{Heinrich, M.P.}, \bibinfo{author}{Liebrand, M.}, \bibinfo{author}{Winzeck, S.}, \bibinfo{author}{Basit, A.}, \bibinfo{author}{Bentley, P.}, \bibinfo{author}{Chen, L.}, \bibinfo{author}{Christiaens, D.}, \bibinfo{author}{Dutil, F.}, \bibinfo{author}{Egger, K.}, \bibinfo{author}{Feng, C.}, \bibinfo{author}{Glocker, B.}, \bibinfo{author}{Götz, M.}, \bibinfo{author}{Haeck, T.}, \bibinfo{author}{Halme, H.L.}, \bibinfo{author}{Havaei, M.}, \bibinfo{author}{Iftekharuddin, K.M.}, \bibinfo{author}{Jodoin, P.M.}, \bibinfo{author}{Kamnitsas, K.}, \bibinfo{author}{Kellner, E.}, \bibinfo{author}{Korvenoja, A.}, \bibinfo{author}{Larochelle, H.}, \bibinfo{author}{Ledig, C.}, \bibinfo{author}{Lee, J.H.}, \bibinfo{author}{Maes, F.}, \bibinfo{author}{Mahmood, Q.}, \bibinfo{author}{Maier-Hein, K.H.}, \bibinfo{author}{McKinley, R.}, \bibinfo{author}{Muschelli, J.},
  \bibinfo{author}{Pal, C.}, \bibinfo{author}{Pei, L.}, \bibinfo{author}{Rangarajan, J.R.}, \bibinfo{author}{Reza, S.M.}, \bibinfo{author}{Robben, D.}, \bibinfo{author}{Rueckert, D.}, \bibinfo{author}{Salli, E.}, \bibinfo{author}{Suetens, P.}, \bibinfo{author}{Wang, C.W.}, \bibinfo{author}{Wilms, M.}, \bibinfo{author}{Kirschke, J.S.}, \bibinfo{author}{Krämer, U.M.}, \bibinfo{author}{Münte, T.F.}, \bibinfo{author}{Schramm, P.}, \bibinfo{author}{Wiest, R.}, \bibinfo{author}{Handels, H.}, \bibinfo{author}{Reyes, M.}, \bibinfo{year}{2017}.
\newblock \bibinfo{title}{Isles 2015 - a public evaluation benchmark for ischemic stroke lesion segmentation from multispectral mri}.
\newblock \bibinfo{journal}{Medical Image Analysis} \bibinfo{volume}{35}, \bibinfo{pages}{250--269}.
\newblock \URLprefix \url{https://www.sciencedirect.com/science/article/pii/S1361841516301268}, \DOIprefix\doi{https://doi.org/10.1016/j.media.2016.07.009}.
\bibitem[{Mehrtash et~al.(2020)Mehrtash, Wells, Tempany, Abolmaesumi and Kapur}]{mehrtash2020confidence}
\bibinfo{author}{Mehrtash, A.}, \bibinfo{author}{Wells, W.M.}, \bibinfo{author}{Tempany, C.M.}, \bibinfo{author}{Abolmaesumi, P.}, \bibinfo{author}{Kapur, T.}, \bibinfo{year}{2020}.
\newblock \bibinfo{title}{Confidence calibration and predictive uncertainty estimation for deep medical image segmentation}.
\newblock \bibinfo{journal}{IEEE transactions on medical imaging} \bibinfo{volume}{39}, \bibinfo{pages}{3868--3878}.
\bibitem[{Mendrik et~al.(2015a)Mendrik, Vincken, Kuijf, Breeuwer, Bouvy, De~Bresser, Alansary, De~Bruijne, Carass, El-Baz et~al.}]{mrbrains_dataset}
\bibinfo{author}{Mendrik, A.M.}, \bibinfo{author}{Vincken, K.L.}, \bibinfo{author}{Kuijf, H.J.}, \bibinfo{author}{Breeuwer, M.}, \bibinfo{author}{Bouvy, W.H.}, \bibinfo{author}{De~Bresser, J.}, \bibinfo{author}{Alansary, A.}, \bibinfo{author}{De~Bruijne, M.}, \bibinfo{author}{Carass, A.}, \bibinfo{author}{El-Baz, A.}, et~al., \bibinfo{year}{2015}a.
\newblock \bibinfo{title}{Mrbrains challenge: online evaluation framework for brain image segmentation in 3t mri scans}.
\newblock \bibinfo{journal}{Comput. Intell. Neurosci.} \bibinfo{volume}{2015}, \bibinfo{pages}{1}.
\bibitem[{Mendrik et~al.(2015b)Mendrik, Vincken, Kuijf, Breeuwer, Bouvy, De~Bresser, Alansary, De~Bruijne, Carass, El-Baz et~al.}]{mendrik2015mrbrains}
\bibinfo{author}{Mendrik, A.M.}, \bibinfo{author}{Vincken, K.L.}, \bibinfo{author}{Kuijf, H.J.}, \bibinfo{author}{Breeuwer, M.}, \bibinfo{author}{Bouvy, W.H.}, \bibinfo{author}{De~Bresser, J.}, \bibinfo{author}{Alansary, A.}, \bibinfo{author}{De~Bruijne, M.}, \bibinfo{author}{Carass, A.}, \bibinfo{author}{El-Baz, A.}, et~al., \bibinfo{year}{2015}b.
\newblock \bibinfo{title}{Mrbrains challenge: online evaluation framework for brain image segmentation in 3t mri scans}.
\newblock \bibinfo{journal}{Computational intelligence and neuroscience} \bibinfo{volume}{2015}.
\bibitem[{Menze et~al.(2015)Menze, Jakab, Bauer, Kalpathy-Cramer, Farahani, Kirby, Burren, Porz, Slotboom, Wiest, Lanczi, Gerstner, Weber, Arbel, Avants, Ayache, Buendia, Collins, Cordier, Corso, Criminisi, Das, Delingette, Demiralp, Durst, Dojat, Doyle, Festa, Forbes, Geremia, Glocker, Golland, Guo, Hamamci, Iftekharuddin, Jena, John, Konukoglu, Lashkari, Mariz, Meier, Pereira, Precup, Price, Raviv, Reza, Ryan, Sarikaya, Schwartz, Shin, Shotton, Silva, Sousa, Subbanna, Szekely, Taylor, Thomas, Tustison, Unal, Vasseur, Wintermark, Ye, Zhao, Zhao, Zikic, Prastawa, Reyes and Van~Leemput}]{Menze2015TheBRATSJ}
\bibinfo{author}{Menze, B.H.}, \bibinfo{author}{Jakab, A.}, \bibinfo{author}{Bauer, S.}, \bibinfo{author}{Kalpathy-Cramer, J.}, \bibinfo{author}{Farahani, K.}, \bibinfo{author}{Kirby, J.}, \bibinfo{author}{Burren, Y.}, \bibinfo{author}{Porz, N.}, \bibinfo{author}{Slotboom, J.}, \bibinfo{author}{Wiest, R.}, \bibinfo{author}{Lanczi, L.}, \bibinfo{author}{Gerstner, E.}, \bibinfo{author}{Weber, M.A.}, \bibinfo{author}{Arbel, T.}, \bibinfo{author}{Avants, B.B.}, \bibinfo{author}{Ayache, N.}, \bibinfo{author}{Buendia, P.}, \bibinfo{author}{Collins, D.L.}, \bibinfo{author}{Cordier, N.}, \bibinfo{author}{Corso, J.J.}, \bibinfo{author}{Criminisi, A.}, \bibinfo{author}{Das, T.}, \bibinfo{author}{Delingette, H.}, \bibinfo{author}{Demiralp, C.}, \bibinfo{author}{Durst, C.R.}, \bibinfo{author}{Dojat, M.}, \bibinfo{author}{Doyle, S.}, \bibinfo{author}{Festa, J.}, \bibinfo{author}{Forbes, F.}, \bibinfo{author}{Geremia, E.}, \bibinfo{author}{Glocker, B.}, \bibinfo{author}{Golland, P.}, \bibinfo{author}{Guo, X.},
  \bibinfo{author}{Hamamci, A.}, \bibinfo{author}{Iftekharuddin, K.M.}, \bibinfo{author}{Jena, R.}, \bibinfo{author}{John, N.M.}, \bibinfo{author}{Konukoglu, E.}, \bibinfo{author}{Lashkari, D.}, \bibinfo{author}{Mariz, J.A.}, \bibinfo{author}{Meier, R.}, \bibinfo{author}{Pereira, S.}, \bibinfo{author}{Precup, D.}, \bibinfo{author}{Price, S.J.}, \bibinfo{author}{Raviv, T.R.}, \bibinfo{author}{Reza, S.M.S.}, \bibinfo{author}{Ryan, M.}, \bibinfo{author}{Sarikaya, D.}, \bibinfo{author}{Schwartz, L.}, \bibinfo{author}{Shin, H.C.}, \bibinfo{author}{Shotton, J.}, \bibinfo{author}{Silva, C.A.}, \bibinfo{author}{Sousa, N.}, \bibinfo{author}{Subbanna, N.K.}, \bibinfo{author}{Szekely, G.}, \bibinfo{author}{Taylor, T.J.}, \bibinfo{author}{Thomas, O.M.}, \bibinfo{author}{Tustison, N.J.}, \bibinfo{author}{Unal, G.}, \bibinfo{author}{Vasseur, F.}, \bibinfo{author}{Wintermark, M.}, \bibinfo{author}{Ye, D.H.}, \bibinfo{author}{Zhao, L.}, \bibinfo{author}{Zhao, B.}, \bibinfo{author}{Zikic, D.}, \bibinfo{author}{Prastawa, M.},
  \bibinfo{author}{Reyes, M.}, \bibinfo{author}{Van~Leemput, K.}, \bibinfo{year}{2015}.
\newblock \bibinfo{title}{The multimodal brain tumor image segmentation benchmark (brats)}.
\newblock \bibinfo{journal}{IEEE Transactions on Medical Imaging} \bibinfo{volume}{34}, \bibinfo{pages}{1993--2024}.
\newblock \DOIprefix\doi{10.1109/TMI.2014.2377694}.
\bibitem[{Minderer et~al.(2021)}]{minderer2021revisiting}
\bibinfo{author}{Minderer}, et~al., \bibinfo{year}{2021}.
\newblock \bibinfo{title}{Revisiting the calibration of modern neural networks}, in: \bibinfo{booktitle}{NeurIPS}.
\bibitem[{Mukhoti et~al.(2020)Mukhoti, Kulharia, Sanyal, Golodetz, Torr and Dokania}]{mukhoti2020calibrating}
\bibinfo{author}{Mukhoti, J.}, \bibinfo{author}{Kulharia, V.}, \bibinfo{author}{Sanyal, A.}, \bibinfo{author}{Golodetz, S.}, \bibinfo{author}{Torr, P.H.}, \bibinfo{author}{Dokania, P.K.}, \bibinfo{year}{2020}.
\newblock \bibinfo{title}{Calibrating deep neural networks using focal loss}, in: \bibinfo{booktitle}{NeurIPS}.
\bibitem[{M{\"u}ller et~al.(2019)M{\"u}ller, Kornblith and Hinton}]{muller2019does}
\bibinfo{author}{M{\"u}ller, R.}, \bibinfo{author}{Kornblith, S.}, \bibinfo{author}{Hinton, G.}, \bibinfo{year}{2019}.
\newblock \bibinfo{title}{When does label smoothing help?}, in: \bibinfo{booktitle}{NeurIPS}.
\bibitem[{Naeini et~al.(2015)Naeini, Cooper and Hauskrecht}]{naeini2015obtaining}
\bibinfo{author}{Naeini, M.P.}, \bibinfo{author}{Cooper, G.}, \bibinfo{author}{Hauskrecht, M.}, \bibinfo{year}{2015}.
\newblock \bibinfo{title}{Obtaining well calibrated probabilities using bayesian binning}, in: \bibinfo{booktitle}{Twenty-Ninth AAAI Conference on Artificial Intelligence}.
\bibitem[{Niculescu-Mizil and Caruana(2005)}]{niculescu2005predicting}
\bibinfo{author}{Niculescu-Mizil, A.}, \bibinfo{author}{Caruana, R.}, \bibinfo{year}{2005}.
\newblock \bibinfo{title}{Predicting good probabilities with supervised learning}, in: \bibinfo{booktitle}{Proceedings of the 22nd international conference on Machine learning}, pp. \bibinfo{pages}{625--632}.
\bibitem[{Oktay et~al.(2018)Oktay, Schlemper, Folgoc, Lee, Heinrich, Misawa, Mori, McDonagh, Hammerla, Kainz et~al.}]{attunet}
\bibinfo{author}{Oktay, O.}, \bibinfo{author}{Schlemper, J.}, \bibinfo{author}{Folgoc, L.L.}, \bibinfo{author}{Lee, M.}, \bibinfo{author}{Heinrich, M.}, \bibinfo{author}{Misawa, K.}, \bibinfo{author}{Mori, K.}, \bibinfo{author}{McDonagh, S.}, \bibinfo{author}{Hammerla, N.Y.}, \bibinfo{author}{Kainz, B.}, et~al., \bibinfo{year}{2018}.
\newblock \bibinfo{title}{Attention u-net: Learning where to look for the pancreas}.
\newblock \bibinfo{journal}{arXiv preprint arXiv:1804.03999} .
\bibitem[{Ovadia et~al.(2019)Ovadia, Fertig, Ren, Nado, Sculley, Nowozin, Dillon, Lakshminarayanan and Snoek}]{ovadia2019can}
\bibinfo{author}{Ovadia, Y.}, \bibinfo{author}{Fertig, E.}, \bibinfo{author}{Ren, J.}, \bibinfo{author}{Nado, Z.}, \bibinfo{author}{Sculley, D.}, \bibinfo{author}{Nowozin, S.}, \bibinfo{author}{Dillon, J.V.}, \bibinfo{author}{Lakshminarayanan, B.}, \bibinfo{author}{Snoek, J.}, \bibinfo{year}{2019}.
\newblock \bibinfo{title}{Can you trust your model's uncertainty? evaluating predictive uncertainty under dataset shift}, in: \bibinfo{booktitle}{NeurIPS}.
\bibitem[{Pereyra et~al.(2017)Pereyra, Tucker, Chorowski, Kaiser and Hinton}]{pereyra2017regularizing}
\bibinfo{author}{Pereyra, G.}, \bibinfo{author}{Tucker, G.}, \bibinfo{author}{Chorowski, J.}, \bibinfo{author}{Kaiser, {\L}.}, \bibinfo{author}{Hinton, G.}, \bibinfo{year}{2017}.
\newblock \bibinfo{title}{Regularizing neural networks by penalizing confident output distributions}, in: \bibinfo{booktitle}{ICLR}.
\bibitem[{Platt et~al.(1999)}]{platt1999probabilistic}
\bibinfo{author}{Platt, J.}, et~al., \bibinfo{year}{1999}.
\newblock \bibinfo{title}{Probabilistic outputs for support vector machines and comparisons to regularized likelihood methods}.
\newblock \bibinfo{journal}{Advances in large margin classifiers} \bibinfo{volume}{10}, \bibinfo{pages}{61--74}.
\bibitem[{Ronneberger et~al.(2015)Ronneberger, Fischer and Brox}]{unet}
\bibinfo{author}{Ronneberger, O.}, \bibinfo{author}{Fischer, P.}, \bibinfo{author}{Brox, T.}, \bibinfo{year}{2015}.
\newblock \bibinfo{title}{{U-Net: Convolutional Networks for Biomedical Image Segmentation}}, in: \bibinfo{booktitle}{Medical Image Computing and Computer-Assisted Intervention -- MICCAI 2015}, pp. \bibinfo{pages}{234--241}.
\bibitem[{Szegedy et~al.(2016)Szegedy, Vanhoucke, Ioffe, Shlens and Wojna}]{szegedy2016rethinking}
\bibinfo{author}{Szegedy, C.}, \bibinfo{author}{Vanhoucke, V.}, \bibinfo{author}{Ioffe, S.}, \bibinfo{author}{Shlens, J.}, \bibinfo{author}{Wojna, Z.}, \bibinfo{year}{2016}.
\newblock \bibinfo{title}{Rethinking the inception architecture for computer vision}, in: \bibinfo{booktitle}{CVPR}.
\bibitem[{Tomani et~al.(2021)Tomani, Gruber, Erdem, Cremers and Buettner}]{Tomani2021Posthoc}
\bibinfo{author}{Tomani, C.}, \bibinfo{author}{Gruber, S.}, \bibinfo{author}{Erdem, M.E.}, \bibinfo{author}{Cremers, D.}, \bibinfo{author}{Buettner, F.}, \bibinfo{year}{2021}.
\newblock \bibinfo{title}{Post-hoc uncertainty calibration for domain drift scenarios}, in: \bibinfo{booktitle}{CVPR}.
\bibitem[{Wang et~al.(2019)Wang, Li, Aertsen, Deprest, Ourselin and Vercauteren}]{wang2019aleatoric}
\bibinfo{author}{Wang, G.}, \bibinfo{author}{Li, W.}, \bibinfo{author}{Aertsen, M.}, \bibinfo{author}{Deprest, J.}, \bibinfo{author}{Ourselin, S.}, \bibinfo{author}{Vercauteren, T.}, \bibinfo{year}{2019}.
\newblock \bibinfo{title}{Aleatoric uncertainty estimation with test-time augmentation for medical image segmentation with convolutional neural networks}.
\newblock \bibinfo{journal}{Neurocomputing} \bibinfo{volume}{338}, \bibinfo{pages}{34--45}.
\bibitem[{Wenzel et~al.(2020)Wenzel, Snoek, Tran and Jenatton}]{wenzel2020hyperparameter}
\bibinfo{author}{Wenzel, F.}, \bibinfo{author}{Snoek, J.}, \bibinfo{author}{Tran, D.}, \bibinfo{author}{Jenatton, R.}, \bibinfo{year}{2020}.
\newblock \bibinfo{title}{Hyperparameter ensembles for robustness and uncertainty quantification}, in: \bibinfo{booktitle}{NeurIPS}.
\bibitem[{Xie et~al.(2016)Xie, Wang, Wei, Wang and Tian}]{xie2016disturblabel}
\bibinfo{author}{Xie, L.}, \bibinfo{author}{Wang, J.}, \bibinfo{author}{Wei, Z.}, \bibinfo{author}{Wang, M.}, \bibinfo{author}{Tian, Q.}, \bibinfo{year}{2016}.
\newblock \bibinfo{title}{Disturblabel: Regularizing cnn on the loss layer}, in: \bibinfo{booktitle}{CVPR}.
\bibitem[{Zhang et~al.(2020)Zhang, Kailkhura and Han}]{zhang2020mix}
\bibinfo{author}{Zhang, J.}, \bibinfo{author}{Kailkhura, B.}, \bibinfo{author}{Han, T.}, \bibinfo{year}{2020}.
\newblock \bibinfo{title}{Mix-n-match: Ensemble and compositional methods for uncertainty calibration in deep learning}, in: \bibinfo{booktitle}{ICML}.
\bibitem[{Zhang et~al.(2019)Zhang, Dalca and Sabuncu}]{zhang2019confidence}
\bibinfo{author}{Zhang, Z.}, \bibinfo{author}{Dalca, A.V.}, \bibinfo{author}{Sabuncu, M.R.}, \bibinfo{year}{2019}.
\newblock \bibinfo{title}{Confidence calibration for convolutional neural networks using structured dropout}.
\newblock \bibinfo{journal}{arXiv preprint arXiv:1906.09551} .
\bibitem[{Zhou et~al.(2020)Zhou, Siddiquee, Tajbakhsh and Liang}]{unetpp}
\bibinfo{author}{Zhou, Z.}, \bibinfo{author}{Siddiquee, M.M.R.}, \bibinfo{author}{Tajbakhsh, N.}, \bibinfo{author}{Liang, J.}, \bibinfo{year}{2020}.
\newblock \bibinfo{title}{Unet++: Redesigning skip connections to exploit multiscale features in image segmentation}.
\newblock \bibinfo{journal}{IEEE Transactions on Medical Imaging} \bibinfo{volume}{39}, \bibinfo{pages}{1856--1867}.
\newblock \DOIprefix\doi{10.1109/TMI.2019.2959609}.

\end{thebibliography}
